**Corresponding author:**
Willem van der Maden;
Email: willem.maden@gmail.com

# Developing and evaluating a design method for positive artificial intelligence

Willem van der Maden , Derek Lomas and Paul Hekkert

Department of Human-centered Design, Faculty of Industrial Design Engineering, Delft University of Technology, Delft, Netherlands

## Abstract

In an era where artificial intelligence (AI) permeates every facet of our lives, the imperative to steer AI development toward enhancing human wellbeing has never been more critical. However, the development of such *positive AI* poses substantial challenges due to the current lack of mature methods for addressing the complexities that designing AI for wellbeing poses. This article presents and evaluates the positive AI design method aimed at addressing this gap. The method provides a human-centered process for translating wellbeing aspirations into concrete interventions. First, we explain the method's key steps: (1) contextualizing, (2) operationalizing, (3) designing, and (4) implementing supported by (5) continuous measurement for iterative feedback cycles. We then present a multi-case study where novice designers applied the method, revealing strengths and weaknesses related to efficacy and usability. Next, an expert evaluation study assessed the quality of the case studies' outcomes, rating them moderately high for feasibility, desirability, and plausibility of achieving intended wellbeing benefits. Together, these studies provide preliminary validation of the method's ability to improve AI design, while identifying opportunities for enhancement. Building on these insights, we propose adaptations for future iterations of the method, such as the inclusion of wellbeing-related heuristics, suggesting promising avenues for future work. This human-centered approach shows promise for realizing a vision of "AI for wellbeing" that does not just avoid harm, but actively promotes human flourishing.

## Introduction

It is 3 a.m., and the familiar prompt flashes across my screen: "Are you still watching?" The question jolts me back to reality. A wave of regret washes over me as I look at the time, knowing the early morning ahead. Yet here I am, lost in another late-night binge session. I cannot help but wonder – what if this time spent indulging my streaming habits could have somehow contributed to my wellbeing instead of harming it?

Picture a streaming service that goes beyond providing entertainment and helps cultivate meaningful social connections. Or imagine a dating app designed to foster more than superficial hook-ups – one that nurtures emotional intelligence and healthy relationship skills with every swipe. Such systems, driven by artificial intelligence (AI), may seem idealistic. However, as AI becomes increasingly integrated into society, the demand for systems that are socially beneficial and promote human flourishing grows (Tomašev et al., 2020; Stahl et al., 2021; Shneiderman, 2022; Nature Human Behavior, 2023).

The impact of a technology often originates from the values inherent in its design (Crawford, 2021; Klenk, 2021). As such, the values manifested in an AI system are the result of deliberate choices made during its design process (Fokkinga et al., 2020; van de Poel, 2020). Recognizing this, there is an emerging opportunity for establishing consensus on methodologies that purposefully integrate these values into the design of AI-driven systems (Morley et al., 2020).

In this article, we investigate the development of *AI* through the lens of *positive* design (Desmet and Pohlmeyer, 2013). Consequently, we focus on the design of AI-driven systems that promote wellbeing.[1] Scholars advocate for using wellbeing as a practical guidepost for beneficial AI development, as it offers an empirically grounded, outcome-focused approach rooted in people's lived experiences (Musikanski et al., 2020; Schiff et al., 2020; Shahriari and Shahriari, 2017; Stray, 2020). Specifically, wellbeing frameworks compile multidimensional metrics that translate abstract principles into measurable indicators grounded in social science.

While the need for positive AI is clear, how to achieve it remains an open question. To address this gap, we present the development and evaluation of a "positive AI design method" that

---

[1.]While conceptualizing wellbeing is one of the challenges this method seeks to address, it can broadly be understood as "experiences of pleasure and purpose over time" (Dolan, 2014, p. 39). However, we draw from the third wave of positive psychology, which means that the method is attuned to the complexities and varied contextual factors that shape wellbeing (Lomas et al., 2021).

integrates insights from positive design (Desmet and Pohlmeyer, 2013), positive computing (Calvo and Peters, 2014), human-centered design (Norman, 2005; Giacomin, 2014; Boy, 2017), and cybernetics (Dobbe et al., 2021; Glanville, 2014; Martelaro and Ju, 2018; Sweeting, 2016) to develop AI for wellbeing. Efforts to integrate ethical values into AI design, such as value-sensitive design (VSD), have been recognized for their potential to align AI systems with broader societal values (Umbrello and van de Poel, 2021). However, these approaches often fall short in providing mechanisms to verify whether the intended values are genuinely realized in AI design outcomes (Sadek et al., 2023c). To effectively design AI for wellbeing, however, it is imperative to rigorously assess its real-world impact (Peters et al., 2020). Building on existing efforts, we investigate how the assessment of AI's wellbeing impact may enhance design approaches, developing a method that proactively integrates wellbeing (assessment) as a core objective of AI design.

This article is primarily aimed at designers seeking to deepen their engagement with the field of AI and AI practitioners, defined in a broad sense, who are interested in designing AI systems that promote wellbeing. Through the lens of positive design, we explore methodologies and frameworks that can bridge the gap between AI technology and human-centered design, offering insights and practical guidance for these audiences. By adopting the cybernetic perspective, we centralize the assessment of wellbeing impact within the AI design process. Our core objective is to evaluate the credibility and robustness of the positive AI design method. To achieve this, we will follow the framework proposed by Cash et al. (2023), presenting a "chain of evidence" that supports our approach. In doing so, we aim to answer the following four research questions:

1. How might we standardize a method for designing AI that actively supports wellbeing?
2. What are the strengths and weaknesses of the method in practical applications?
3. To what extent does the method yield successful design outcomes?
4. How can future iterations of the method enhance its credibility and robustness?

The remainder of the study is structured as follows:

- **Background:** It discusses definitions of AI and the need for a human-centered AI design approach, highlighting gaps in current methodologies and addressing *RQ1*.
- **Design method:** It outlines how the method was developed and refined and the key steps of the method, answering *RQ1*.
- **Multi-case study:** It presents case studies of novice designers using the method, showcasing its practical strengths and weaknesses and addressing *RQ2*.
- **Expert evaluation study:** It reports on an expert evaluation of concepts from the positive AI design method, directly relating to *RQ3*.
- **Discussion and future directions:** It discusses limitations and proposes enhancements for the method, reflecting on its contribution to human-centered AI and future research directions, answering *RQ4*.

## Background

In this section, we explore a definition of AI, scoping it as a special type of sociotechnical system through the concept of cybernetics. We further identify the key challenges and opportunities for incorporating human wellbeing into AI systems, setting the stage for the development of the positive AI design method.

### *Artificial Intelligence*

The term "AI" carries a breadth of meanings that have evolved alongside its advancements. The essence of AI, as pointed out by AI pioneer John McCarthy, morphs as its applications become ubiquitous in everyday technology (Vardi, 2012). At the center is the notion of "intelligence" itself. Although definitions vary, they commonly highlight abilities in reasoning, problem-solving, and adapting to new challenges (Sternberg, 2003). In an effort to integrate these recurring themes, AI researchers Legg and Hutter (2007, p. 9) propose defining intelligence as "an agent's ability to achieve goals in a wide range of environments." This perspective suggests that intelligence, fundamentally, is about an entity's adaptability and its proficiency in navigating a spectrum of scenarios to achieve its goals.

The "artificial" aspect of AI lies in its *deliberate design*, contrasting with biological intelligence that naturally occurs in living organisms (Gabriel, 2020). As such, AI research focuses on *building* intelligent agents that choose actions to maximize performance based on received inputs and inherent knowledge, where agents perceive their environment through **sensors** and act upon it through **actuators** (Russell and Norvig, 2022, pp. 54–58).

Building on this understanding of AI, it is essential to recognize that AI systems exist within a complex sociotechnical context. Dobbe et al. (2021) highlight the frequent discrepancy between the promised benefits of AI systems and their actual consequences, termed the "sociotechnical gap." This gap arises from the divergence between socially necessary outcomes and what AI can technically achieve. For example, while a recommender system may aim to provide valuable suggestions to users, in practice, it could inadvertently promote misinformation or polarization.

To address this challenge, various scholars have proposed understanding AI as a sociotechnical system that encompasses not only its technical capabilities but also its limitations and the governance structures surrounding it (Dean et al., 2021; Dobbe et al., 2021; Krippendorff, 2023; Selbst et al., 2019; Stray, 2020; van de Poel, 2020; van der Maden et al., 2022; Vassilakopoulou, 2020). For instance, ChatGPT should be considered in terms of not merely its underlying model but also its user interface, the company behind it, public perceptions, and the various use cases and purposes it serve. As such, some of these scholars advocate for adopting a cybernetic perspective, which emphasizes the importance of feedback and adaptation in managing the inherent complexities of sociotechnical systems, giving rise to the inclusion of non-technical and natural entities (Dobbe et al., 2021; Pangaro, 2021; van der Maden et al., 2022; Krippendorff, 2023).

### *Cybernetics: AI as sociotechnical system*

Cybernetics, which emerged in the 1940s, is a transdisciplinary field focused on communication, control, and circular causality in systems (Mindell, 2000). The term "cybernetics" is derived from the Greek infinitive "kybernao," meaning "to steer, navigate, or govern." A core concept in cybernetics is the feedback loop, which creates circular causality between a system's past, present, and future states (Wiener, 1961). At its core, cybernetics presents an alternative perspective to traditional AI design by emphasizing the symbiotic relationship between humans and machines within complex sociotechnical systems (Mead, 1968; von Foerster, 2003). It focuses on the dynamics of feedback loops, communication, and control mechanisms that underpin both biological and mechanical systems, proposing that understanding these can enhance the design and function of AI (Beer et al., 1990; Sato, 1991).

By viewing AI through a cybernetic lens, designers are encouraged to consider AI not just as isolated algorithms but as part of an interconnected web of social, technological, and environmental factors (Dobbe et al., 2021; Scott, 2004). This perspective offers a holistic framework for understanding the challenges around designing AI across various systemic layers. As such, the design of positive AI interventions can go beyond the algorithm and even the platform itself. For example, companies may use the method to make adaptations to the recommender systems that govern their platforms, while smaller design firms may develop a third-party add-on that alters the interaction with a platform to support wellbeing. Interventions can even take place in the broader ecosystem, such as the development of institutional guidelines for use of ChatGPT in education – thus lowering student anxiety in using these tools while bolstering its potential educational impact.

### Challenges of designing AI

When discussing the design AI systems, what specific process are we referring to? The development of AI is often conceptualized as a multi-stage life cycle, traditionally segmented into seven key stages,[2] with the design phase being integral in translating business cases into engineering requirements (Morley et al., 2020). This phase is critical, as it lays the groundwork for how AI systems will function and interact within human contexts. Norman and Stappers (2015) advocate for the involvement of designers throughout the entire development process of sociotechnical systems. However, this article primarily addresses the design phase. This emphasis does not detract from the importance of a holistic approach – which we strongly support – but rather aims to provide a detailed examination of the unique challenges and opportunities within this specific phase. By concentrating on the design phase, we aim to address the nuances that shape the early and critical decisions in AI development, understanding that these decisions have far-reaching implications for all subsequent stages.

Returning to the question at hand, Yang et al. (2020) highlight that designing effective AI systems presents unique challenges distinct from those encountered with traditional software systems. They argue that the uncertainty surrounding AI capabilities and complexity of possible outputs makes it difficult to ideate, prototype, and evaluate human–AI interaction using standard HCI methods. As they point out, AI systems continue adapting after deployment, so designers struggle to anticipate changing behaviors across contexts. Additionally, the near-infinite output possibilities, especially for adaptive AI, mean that traditional prototyping fails to capture the full range of behaviors and experiences.

Furthermore, as we will address later, effectively incorporating wellbeing into AI design demands engaging with user communities. However, integrating user communities into the AI development process is challenging because of technical complexities, the unpredictable evolution of AI technologies (Sadek et al., 2023b), significant communication gaps (Piorkowski et al., 2021), and lack of relevant expertise (Hsu et al., 2022). These difficulties are exacerbated when designing for values such as wellbeing, as they are complex, multifaceted (Schwartz et al., 2012), and interpreted differently across individuals (Graham et al., 2009) and cultures (Sachdeva et al., 2011). As such, there are many possible interpretations of values like fairness, trustworthiness, and empathy, as well as disagreement over their relative importance (Jakesch et al., 2022).

While there is broad aspiration toward high-level AI ethics principles like fairness and transparency, translating these into practice remains challenging (Morley et al., 2021; Schiff et al. (2021b). For example, a review of guidelines on AI ethics found extensive discussion of principles like transparency and fairness, but very little on technical explanations for achieving them (Hagendorff, 2020). Similarly, Schiff et al. (2021a) underscore the complexity of applying ethical principles like fairness and transparency across sectors, highlighting a gap in consensus on practical implementation, which directly impacts the integration of values such as wellbeing into AI systems. Bridging this divide between principles and practice remains an open research challenge. It requires developing methods that reduce the indeterminacy of abstract norms while retaining adaptability to diverse contexts (Jacobs and Huldtgren, 2021).

In this regard, we may look to VSD as a promising methodology for embedding abstract values such as privacy into concrete design specifications, thereby guiding AI systems to better serve and reflect the diverse needs of stakeholders while promoting inclusivity and human-centricity in technology (Zhu et al., 2018b). For instance, Umbrello and van de Poel (2021) present a case study in which they successfully translated crucial values like non-maleficence into actionable design criteria for a novel AI system.

However, Sadek et al. (2023b) note that a significant shortfall in current VSD practices is their inability to effectively assess whether these values are genuinely reflected in the outcomes of AI systems, highlighting a lack of impact assessment mechanisms (both qualitative and quantitative). It is this gap that our method tries to fill for two reasons. First, for any impact-centered method (which arguably any value-oriented design project is), it is essential to establish causal links between interventions and system fluctuations (Fokkinga et al., 2020) – mere good intentions do not cut it. Second, as Schiff et al. (2020) point out, impact measurement leads to evidence-based decision-making and promotes accountability, thus fostering iterative improvement. Now that we have an overview of the challenges related to designing AI, let us turn our attention to the additional challenges introduced by a focus on wellbeing.

### Challenges of designing AI for wellbeing

Designing AI systems specifically to enhance human wellbeing introduces additional complexities. That is, wellbeing is inherently multifaceted, is variable across individuals, and manifests differently across cultural contexts, making it difficult to define and design for in a measurable way (Halleröd and Seldén, 2013; Huppert, 2017). AI systems often optimize narrow objectives, making it hard to ensure that they improve wellbeing holistically. Instead of promoting broad human flourishing, they tend to target limited metrics. This narrow focus in AI systems' design and implementation is recognized as one of the six grand challenges for human-centered AI (Ozmen Garibay et al., 2023). To address these challenges further, a recent article identified seven key challenges for designing AI *for wellbeing* specifically (van der Maden et al., 2023b). They used a cybernetic framework to group the challenges into four categories, listed below and mapped to a simple schematic in Figure 1.

---

[2]A complete AI development life cycle includes seven stages: business and use-case development, design phase, data procurement, building, testing, deployment, and monitoring (Morley et al., 2020).

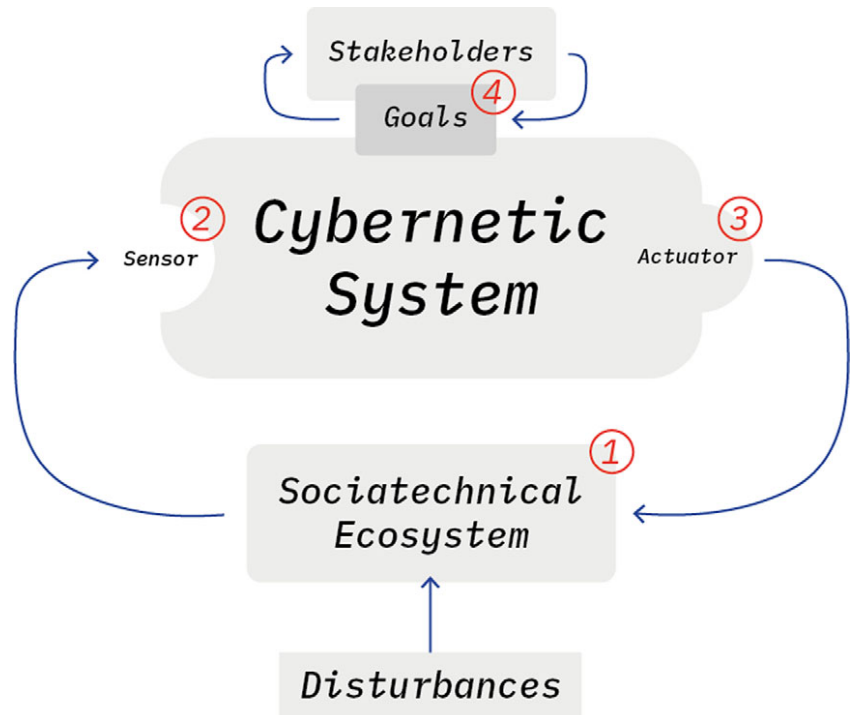

**Figure 1.** It shows a schematic representation of a cybernetic system. The different categories of challenges can be mapped onto this framework: (1) understanding the system context which entails modeling the relation between wellbeing of the systems constituents and its various components; (2) operationalizing said model of wellbeing; (3) designing interventions to actively promote operationalized model of wellbeing; and (4) retaining alignment with the overall goal. The latter refers to both challenges of algorithmic optimization and scrutinizing the objective (e.g., Is the wellbeing objective still aligned to needs and desires of all relevant stakeholders?) Used with permission from van der Maden et al., (2023b).

- **Conceptualization of wellbeing:** the challenges around choosing the appropriate theoretical paradigm for conceptualizing wellbeing and modeling wellbeing contextually given its complexity and unclear relationships between system components and wellbeing facets.
- **Operationalization of wellbeing:** the challenges of measuring wellbeing contextually with adaptive instruments, translating small-scale qualitative wellbeing data to large-scale metrics suitable for optimization algorithms, correlating self-report and behavioral data collection, and reconciling the different paces at which wellbeing changes versus AI optimization occurs.
- **Designing AI actions that promote wellbeing:** the lack of mature methods and examples for putting wellbeing at the core of AI system design, beyond just avoiding harm.

Most tools focus on alignment but lack concrete guidance on promoting human flourishing.

- **Optimizing for wellbeing:** the challenges of making trade-offs between competing objectives (e.g., individual versus communal wellbeing) when optimizing AI systems for wellbeing, and dealing with the fundamental constraint that wellbeing changes slowly while AI optimization is rapid.

Several frameworks have been developed to guide the design of wellbeing-supportive technology, such as the framework by Wiese et al. (2020) that maps wellbeing-enhancing activities (Lyubomirsky and Layous, 2013) to digital technology design and the "METUX" framework by Peters et al. (2018) that supports wellbeing in digital experiences. However, these methods were not specifically developed for AI and do not directly address all four challenges mentioned earlier. The IEEE-7010 standard (Schiff et al., 2020), however, provides a more holistic approach tailored for the development and assessment of autonomous and intelligent systems (A/IS) with human wellbeing at the forefront. This standard offers an iterative wellbeing impact assessment (WIA) process, stakeholder engagement, and a focus on wellbeing indicators across various domains. Through these facets, the standard addresses three of the four challenges by supporting the conceptualization, operationalization, and optimization of wellbeing. However, as a standard rather than a design method, it does not directly guide designers in translating insights from the assessment into concrete design interventions.

Thus, our goal is to develop a method that not only builds upon existing frameworks, such as those outlined in the IEEE-7010 standard, but also harmonizes with recognized design and innovation approaches, including design thinking (Dorst, 2011; Cross, 2023) and the double diamond model (Design Council, 2007). By doing so, we hope to invite human-centered designers to the field of AI and bring human-centered design principles to the development of AI systems.

## A design method for positive AI

The positive AI method is intended to provide designers with a structured process for developing AI systems that actively promote human wellbeing. It aims to address key challenges in conceptualizing, measuring, and designing wellbeing-supportive functionality into AI. It focuses on AI systems that people interact with daily, including curatorial AI (e.g., Tinder, Netflix, and Spotify), generative AI (e.g., ChatGPT), voice assistants (e.g., Alexa), among others. It is important to note that some existing systems do address wellbeing, either intentionally or inadvertently. For example, Stray (2020) points out that Facebook and Google have made deliberate efforts to support wellbeing through their platforms. However, the positive AI method aims to go further by ensuring that the integration of wellbeing is an active goal from the outset of system design, rather than an afterthought. This includes active integration in existing platforms and applications (e.g., Instagram, Claude, and YouTube) as well as dedicated[3] integration in novel platforms and applications (e.g., Hume's Empathic Voice Interface and Headspace).[4]

By empowering designers and AI practitioners with concrete techniques, the method aims to create AI that measurably improves human thriving. It represents an initial attempt to address the lack of practical guidance in existing AI ethics literature specifically regarding enhancing wellbeing (Morley et al., 2020; Schiff et al., 2020).

### Development of the method

The positive AI method was developed using a cybernetic approach as an organizing framework following earlier discussions. Cybernetics views systems as cyclic processes of sensing states, comparing to goals, and taking action (Mindell, 2000). This perspective enabled organizing the design challenges into distinct phases, with each phase addressing a different category of challenges (Fig. 1), while acknowledging the inherent entanglement present in complex sociotechnical systems (Dobbe et al., 2021).

We developed this method following a research-through-design process that drew inspiration from existing frameworks such as the earlier mentioned IEEE-7010 standard (Schiff et al., 2020). The development involved collaboration between designers, researchers, and students over multiple projects. An initial two-year project designing a cybernetic system for institutional wellbeing during COVID-19 informed the first version of the method (van der Maden et al., 2023a), which incorporated elements of the IEEE-7010 process, such as stakeholder engagement and a focus on wellbeing indicators across various domains. This early version was then refined through an iterative process that involved scrutinizing the method's efficacy and incorporating community feedback. The refinement focused on streamlining the method's steps, improving its understandability, and enhancing its relevance. These various versions were then tested in five design courses given at the master's level, where student teams designed AI systems aiming to support wellbeing. These findings were then consolidated to present the version that is evaluated in this article.

Further, the positive AI method is intended to complement and enhance typical design processes. For example, it parallels the empathize, define, ideate, prototype, test, and implement phases of design thinking (Dorst, 2011; Cross, 2023), with a specific focus on wellbeing and AI. Furthermore, the phases of our method align with the convergence and divergence characteristic of the double diamond framework (Design Council, 2007), while also emphasizing the iterative process inherent in most design strategies and frameworks.

Taking these considerations together, we arrived at five phases in the proposed method. This choice mirrors phases in other design cycles and highlights the importance of restarting the cycle, often less evident in typical AI development and psychological research. To emphasize that continuous improvement of the wellbeing model is central to positive AI, we included restarting the cycle as a separate step. The phases correspond to the four categories of challenges (conceptualization, operationalization, design, and optimization), with an additional stage for implementation.[5] We will discuss these phases in depth next.

### Phases of the positive AI method

In short, the method involves ensuring that AI systems are *sensitive* to factors of human wellbeing and *enabled* to support them. Consequently, the five phases should help the designer to understand wellbeing in context (phase 1), to make it measurable (phase 2), to design systems (inter)actions that promote wellbeing (phase 3), to implement the designs (phase 4), and to sustain alignment (phase 5). Figure 2 shows an overview of the methods phases with brief annotations of the content of each phase. A useful checklist with the activities and outcomes of the method can be found in the Appendix.

#### Phase 1 – Contextualize: understanding wellbeing in context

To be able to sense wellbeing, we first need an understanding of what wellbeing is. However, wellbeing is complex, is multifaceted, and manifests differently across contexts, making it difficult to conceptualize. For instance, despite there being overlap, how wellbeing manifests in an educational setting may differ from wellbeing in a healthcare environment. In education, wellbeing may encompass a sense of purpose, self-efficacy, and belonging, while in health care, it may manifest as physical health and effective pain management.

This complexity extends to AI systems and their broader contexts as well. Evidently, different AI systems, such as social media platforms or dating apps, influence wellbeing in different ways. For instance, the former might impact users' sense of social belonging and community engagement, while the latter might influence aspects of wellbeing related to relationships and self-esteem. In essence, wellbeing is shaped by the interaction between the user, their circumstances, and the specific AI system. Therefore, designers first need to understand how wellbeing *manifests* within their specific context – i.e., how the system they are (re)designing relates to the wellbeing of its user community.

A logical starting point is the extensive theoretical literature on wellbeing (e.g., Alexandrova, 2012, 2017; Diener and Michalos, 2009; Cooke et al., 2016). This provides a wealth of information and can aid in the initial coupling of system components to wellbeing dimensions. However, the breadth of this literature can be overwhelming[6] and may not fully apply to the emergent nature of AI contexts (Kross et al., 2021; Stray et al., 2022). Consequently, designers must prioritize which aspects of wellbeing to focus on for their specific project.

To guide designers in prioritizing which aspects of wellbeing to focus on, we can follow the argument of Harris (2010) that they should focus first on the path that empirically contributes the least to suffering and the most to human flourishing.[7] This necessitates an empirical investigation to determine which dimensions of wellbeing are most relevant within the specific context – the low hanging fruit so to say.[8]

---

[3]Calvo and Peters (2014) distinguish between active and dedicated wellbeing integration, noting that active integration into existing platforms presents additional challenges as wellbeing goals must compete with preexisting objectives, such as those related to revenue.

[4]We primarily focus our discussion on ubiquitous AI, specifically AI-driven platforms and generative AI. However, our method may have broader applicability. While every AI technique, use case, and context brings different challenges, the fundamental considerations discussed here apply to any AI application that intends to consider wellbeing impact as a core objective.

[5]This stage faces practical challenges different from those discussed in van der Maden et al. (2023b), but it is essential for connecting wellbeing, design, and AI development. These challenges are discussed in-depth elsewhere (e.g., Ellefsen et al., 2019; Shaw et al., 2019; Chomutare et al., 2022; Ahmadi, 2023; Donovan, 2024).

[6]That is, due to wellbeing's complexity, there may be too many aspects and theories to consider in this stage of the design process.

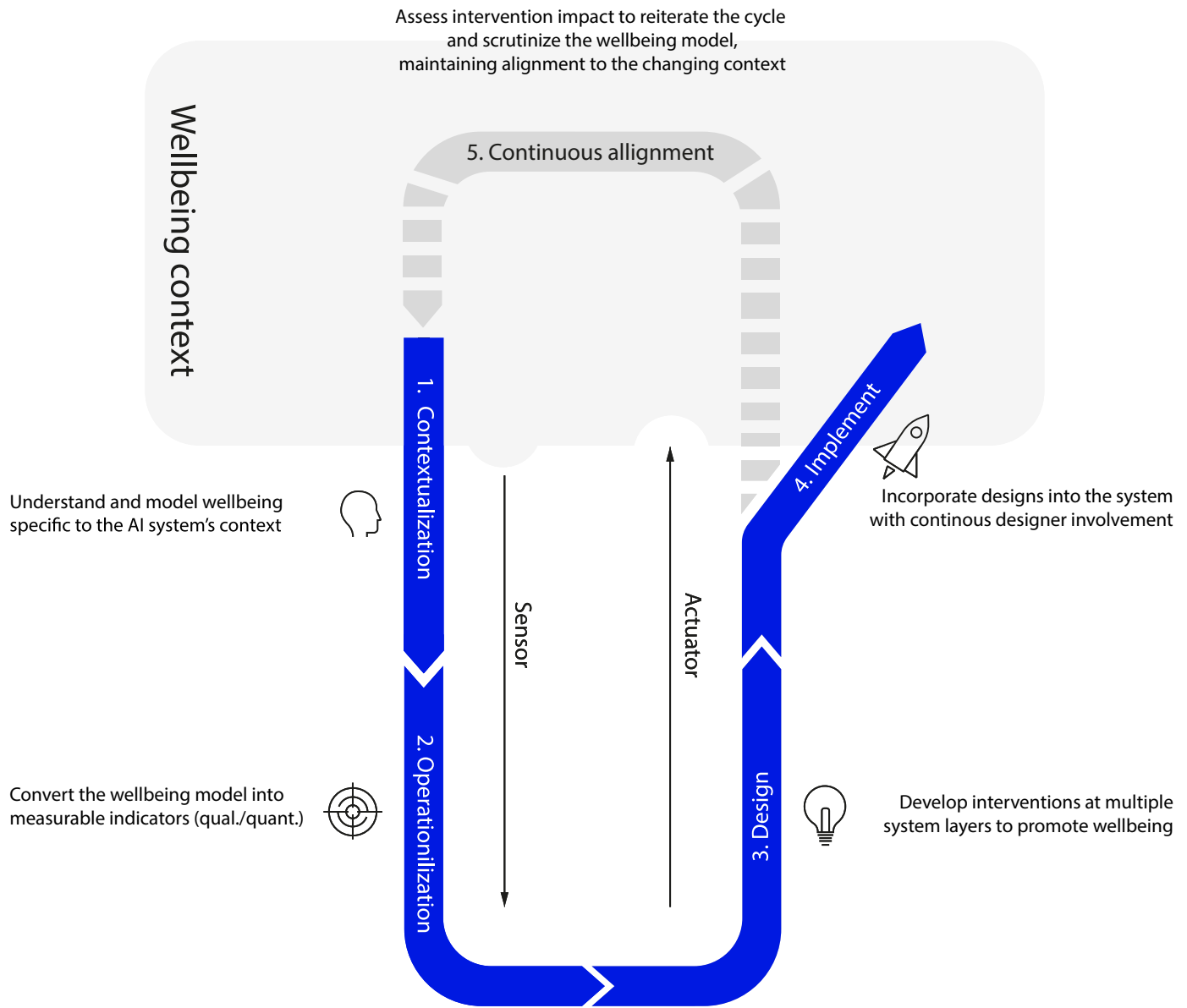

**Figure 2.** Diagram illustrating the positive AI method's cyclical approach within a wellbeing context. Phases 1 (contextualization) and 2 (operationalization) primarily contribute to developing the sensors of the AI system, while phases 3 (design) and 4 (implementation) focus on developing its actuators – reference the section on the background of AI for a discussion of sensors and actuators. The cycle culminates in phase 5 (continuous alignment), demonstrating an ongoing feedback loop between all stages and its environment.

Before being able to engage in such an inquiry, it is essential to develop an initial mapping of the AI system's components. This involves an analysis of the relevant elements within the system, such as its user interface design, underlying algorithms, data processing techniques, and output capabilities. This task is complex and unlikely to be fully resolved in the initial iteration. However, designers have employed a range of techniques to tackle similar challenges, including stakeholder analysis (Friedman et al., 2009), context mapping (Visser et al., 2005), and competitive analysis to gauge the effectiveness of various interventions (Dalpiaz and Parente, 2019). For example, in a social media platform, this would include examining how content algorithms shape user interactions and social norms, while for a dating app, it would involve analyzing the matchmaking algorithms and their impact on user experience and satisfaction. Each profession has a different understanding of what components of a system may yield which effects. Ideally, to capture the full picture of the system's components and their effects, a positive AI team would consist of a multidisciplinary group, including human-centered designers trained to bring together diverse perspectives (Sadek et al., 2023a).

---

[7]Harris' stance should not be interpreted as strictly utilitarian; he does not support maximizing happiness for the majority if it undermines the rights of minorities – a critical issue prevalent in today's AI landscape (Crawford, 2021). However, given the complexity of wellbeing, we must begin our inquiry with a practical choice: opting for paths that minimize suffering and enhance flourishing. There are other ethical frameworks to guide this process such as decolonial, feminist, or care ethics which would likely steer this inquiry in a different direction that warrants further exploration.

[8]The first pass through this contextualization phase is arguably the "hardest" or at least the most "effortful," akin to the cold-start problem in recommender systems. Designers start with minimal data, targeting broad wellbeing aspects. However, as subsequent iterations refine the model, questions sharpen over time, thus deepening the contextual understanding, progressively easing the process, and lowering its resource intensity.

Then, having gained an understanding of the various components of our context, we can engage with the community to develop a more nuanced and detailed understanding of how those parts relate to specific facets of wellbeing. The goal here is to reveal which wellbeing facets seem most influenced or impacted within this particular sociotechnical context. For this, we can employ a slate of human-centered approaches such as interviews, focus groups, and observational studies to identify which manifestations of wellbeing surface as most pronounced and how they relate to the components of the system. However, paraphrasing Alexandrova and Fabian (2022), how can we safeguard high scholarly standards of measurement while opening it up for lay participation? To address this, we recommend grounding this investigation in rigorous methods such as proposed by Layard and De Neve (2023). By oscillating between contextually relevant indicators and scientifically established wellbeing metrics, we can establish a constructive dialogue toward a nuanced yet scholarly model of wellbeing (Loveridge et al., 2020).

In other words, by thoroughly researching the user community and AI ecosystem from both practical and theoretical perspectives, designers can determine which components contribute most to wellbeing in that unique context. This allows designers to strategically focus design efforts on the wellbeing facets that are most relevant and impactful for that specific user community when conceptualizing and designing the AI system.

As a result, at the end of phase 1 (contextualization), designers will have established a contextual model of wellbeing which encompasses hypothesized[9] relations between the most relevant facets of wellbeing and components of the context. This will be achieved by engaging in a conversation between the literature and the context. This dialectic channel can be opened through a plethora of human-centered approaches and is important in facilitating the continuous alignment which we will discuss in phase 5. Before that, the next phase will address how to make the abstract theoretical model measurable.

### Phase 2 – Operationalize: making contextual wellbeing measurable

In this phase, the designer transforms the contextual model from an abstract concept to observable and actionable criteria. These criteria can then be assessed both qualitatively (e.g., interviews, focus groups) and quantitatively (e.g., surveys, experiments). This process is invaluable, not only at the culmination of our design cycle, enabling the assessment of positive AI interventions' impact, but also throughout the design process itself. It allows for ongoing evaluations of whether the prototypes are on track to achieve the desired outcomes and facilitates a continuous scrutiny of our contextual model of wellbeing.

More specifically, by developing observable criteria, we can refine our understanding of wellbeing, uncovering causal relationships, and assess the effectiveness of design interventions. Through this operationalization process, we can empirically investigate the hypothesized connections of our contextual model defined in the previous phase. This mechanic of oscillating between theoretically defining a contextual model and empirically investigating it is core to psychological research into wellbeing (Diener and Michalos, 2009).

Note that it is important to recognize the distinction between measuring a phenomenon like wellbeing and its antecedents or determinants (Blijlevens et al., 2017). That is, we must separate measures of overall wellbeing from context-specific factors that influence it. For example, converting the abstract concept of "social connectedness" in the wellbeing model into a tangible metric might involve measuring both overall social wellbeing through validated scales, as well as specific indicators like the frequency and quality of an individual's social interactions. This allows us to capture the broad construct while also linking it to relevant contextual determinants, thereby revealing potential design spaces. It is in the combination of validated global measures and context-relevant indicators that we find the actionable insights needed to understand and improve wellbeing (van der Maden et al., 2023a).

Quantitative operationalizations of our contextual model are crucial for scaling our investigation and translating insights from local contexts to system-wide applications. For instance, when developing a wellbeing feature for a social media platform, initial tests with a user panel may not fully represent the broader community. To extend these local insights, a survey based on our operationalizations can validate the model's applicability at a system level. Additionally, integrating these operationalizations into AI system optimization processes, including algorithmic adjustments and managerial decisions, can significantly enhance system wellbeing. Operationalized metrics offer local indicators for system performance and wellbeing, facilitating their incorporation into optimization processes for more effective observation and refinement (Stray, 2020).

Finally, this process allows us to assess whether our design interventions produce their intended positive impacts on wellbeing. Such assessments can both be qualitative (e.g., observational studies – does the user engage with our interventions as intended) as well as quantitative (e.g., a controlled experiment comparing the wellbeing scores of two groups over time). This assessment process is essential for complex, interconnected design projects where various elements mutually influence each other (Fokkinga et al., 2020). By introducing interventions in a slow, incremental way, designers are able to couple wellbeing fluctuations to specific system components, hence grounding the positive AI design process in empirical data.

### Phase 3 – Designing: ideating and prototyping

With the contextual model and operationalized wellbeing metrics in hand, designers now have an idea of where in the system they can intervene to achieve specific wellbeing effects. Consider investigating how to enhance ChatGPT's impact: If observation studies and interviews reveal user anxiety about "correct" tool usage – fearing fraud accusations (e.g., Chan and Zhou, 2023) or a sense of lost authorship over produced content (e.g., Amirjalili et al., 2024; van der Maden et al., 2024) – designers may look to ideate ideas to promote user empowerment and authenticity.

Nonetheless, choosing the right design direction can be challenging. To address this, "scaling up the conversation" becomes crucial by verifying the hypothesized relationships identified in phase 1 at a system level. Employing quantitative methods through the operationalizations from phase 2, such as user surveys, behavior tracking, and crowdsourced wellbeing ratings, can help pinpoint areas where interventions may yield the most impact. This approach not only identifies key focus areas but may also generate new ideas from the target audience, as demonstrated in van der Maden et al., (2023a). Should this process not highlight impactful directions, revisiting earlier phases is advisable. It should also be emphasized that adopting a sociotechnical perspective on AI design allows interventions to occur across at least three distinct levels:

- **Experience design** – Crafting the overall user experience arc to positively affect wellbeing trajectories, either with interventions inside or outside the platform (e.g., guidelines for positive use of ChatGPT in education);
- **Interface design** – Leveraging the user interface for wellbeing-promoting interactions (e.g., a pop-up saying "You're all caught up."); and
- **Algorithm design** – Optimizing machine learning and recommendation algorithms to align with wellbeing facets influenced by the system.

---

[9]"Hypothesized" in the sense that we will establish evidence for the causality over the duration of multiple cycles which allows us to assess the relation.

There is no universally optimal design approach for this phase; the choice of technique is influenced by the specific context and scope of the design project. Designers and firms often have preferred methods and are welcome to use these. To effectively kickstart the ideation process, particularly in tackling the previously discussed challenges of designing AI, we recommend two specific resources: the "AI meets Design Toolkit" (Piet and MOBGEN, 2019)[10] and the "AI Design Kit" (Yildirim et al., 2023).[11] These resources stand out for their inclusion of generative prompts designed to aid in conceptualizing machine intelligence features. Both toolkits are instrumental in facilitating a creative and informed ideation process.

At the end of this phase, designers will have produced a range of design strategies and artifacts that translate the operationalized model into actionable interventions aligned with wellbeing goals. The designer may end up with artifacts such as journey maps delineating goal-oriented user flows, wireframes illustrating proposed interfaces, interactive low fidelity prototypes, and explicit design principles encoding wellbeing aims. With these artifacts in hand, the designer then clearly communicates the guiding wellbeing goals and specific envisioned interactions to engineering teams for implementation. Ultimately, the success of this design phase lies in its ability to translate the operationalized model into a resonant yet actionable vision for design interventions that promote wellbeing.

### Phase 4 – Implement: integrating and testing interventions

In this implementation phase, the focus shifts to realizing the conceptualized interventions. This means further developing prototypes and testing them with users, thus putting the designers' vision in effect. In the design of sociotechnical systems, it is important that designers are included in the implementation phase (Norman and Stappers, 2015; Sadek et al., 2023b). It is not solely the domain of development and engineering teams to bring these designs to life; designers must maintain a hands-on presence to guide and refine the implementation.

Specifically, the design artifacts and principles produced in phase 3 provide critical guidance during the implementation phase. In a collaborative effort with these designers, engineers may utilize tangible visions of the system's form and function to construct the necessary components ready for user testing. Additionally, designers refer to these artifacts to steer the ongoing development, ensuring alignment with the wellbeing-centric principles encoded within them. For example, by comparing implemented features with the prototypes and design criteria, designers can identify divergence from the intended interactions. This ability to reference the codified vision facilitates course-correcting implementations back into alignment.

By staying engaged through the implementation phase, designers are better positioned to address any unforeseen challenges that emerge. This proactive approach ensures that the wellbeing impacts, carefully planned in the design phase, are fully realized in the final product. Avoiding shortcuts or efficiency concessions helps maintain the integrity of the project's goals. Making such concessions could potentially compromise the intended outcomes of the approach. The sustained participation of designers is essential in bridging the gap between user needs, technical constraints, and the original design vision. In essence, the artifacts and guiding principles developed in phase 3 play a pivotal role in keeping the implementation firmly anchored to the wellbeing impacts, ensuring that these principles are not lost but rather brought to life in the final integration.

### Phase 5 – Reiterate: sustaining continuous alignment

Finally, maintaining alignment with the system's wellbeing context is crucial. This involves continually assessing whether our interventions meet their intended goals and if the wellbeing model remains applicable. Such evaluations allow us to stay attuned to changes in the wellbeing context and uncover new opportunities for positive intervention. This strategy leverages established communication channels and operates on two levels:

- At the process level, designers should continually engage users and communities during contextualization and design activities. Human-centered methods like interviews, focus groups, and co-design workshops enable aligning design decisions with community goals as they evolve.
- At the system level, implementation marks the end of one iterative cycle. As the loop gets tighter through repeated iterations, the need for major interventions tends to diminish as positive adjustments accumulate. Nonetheless, the designer can step back, evaluate what occurred in relation to the wellbeing model, and determine needs for the next round. Does the contextual model require updating? Were key perspectives missing?

Restarting the loop enables revisiting the contextual understanding and community connections to realign priorities. By continuously iterating alignment at process and system levels, the approach maintains a pulse on emerging wellbeing impacts as user needs and technological capabilities shift. This cycling sustains contextual sensitivity of the wellbeing focus over time.

To effectively implement the positive AI method, team composition and collaboration are crucial. The team should ideally consist of individuals from diverse disciplines, including designers, AI experts, domain specialists, and user representatives. Clear communication channels and protocols facilitate effective collaboration among team members (Morley et al., 2021; Sadek et al., 2023a). Regular meetings, workshops, and documentation help bridge disciplinary gaps and ensure a shared understanding of project goals and progress (Dijk and van der Lugt, 2013). This multidisciplinary approach, combined with strong stakeholder engagement, enables the positive AI method to effectively address real-world challenges and opportunities.

---

[10]Available at https://aixdesign.co/toolkit
[11]Available at https://aixdesign.gumroad.com/l/toolkit

### Method applied to fictional example of a streaming platform

To illustrate the positive AI method in action, this section presents a fictional scenario involving a streaming platform seeking to align with wellbeing goals. In this hypothetical case, the process would begin with a review of literature on video platforms and human–AI interaction to compile a list of key features and hypothesize their impacts on wellbeing. For example, studies suggest personalized video recommendations can sometimes limit users' openness to new perspectives and may create filter bubbles (Ferwerda and Schedl, 2016; Dhelim et al., 2022; Areeb et al., 2023). In contrast, features like custom video playlists may boost users' feelings of autonomy and control over their viewing experiences (Möller et al., 2020). At this stage, our initial theoretical model incorporates hypothesized relationships between specific AI features (such as recommendations and playlists) and key aspects of wellbeing (like openness and autonomy).

Next, we could refine our preliminary conceptual framework through comprehensive user studies. This step would involve conducting interviews, organizing focus groups, and administering surveys. The aim of this research phase is to gather direct feedback on how the platform's functionalities impact user wellbeing, balancing the identification of obstacles with potential improvements. By engaging a diverse participant group, we seek to capture a wide range of insights and user experiences.

We would then operationalize wellbeing by selecting validated scales like the Personal Growth Initiative Scale (Robitschek et al., 2012) to quantify growth and openness to new ideas.

Additionally, we would develop context-specific metrics, such as an aggregated playlist complexity score. This local metric could be calculated from factors such as breadth of topics, diversity of creators, and degree of organizational structure in users' video playlists. It would serve as an indicator of the level of perceived control over viewing experiences.

Equipped with this contextualized model of the platform's wellbeing impacts, we could propose targeted interventions to optimize the scales and metrics. For instance, one could suggest an algorithmic adjustment that sporadically introduces unexpected video recommendations, motivating users to explore content beyond their regular preferences, thereby potentially elevating personal growth metrics.

To implement such proposals, collaborative sessions with designers, engineers, and users would allow iteratively developing and refining features based on observed wellbeing impacts and user feedback. Designers would facilitate participatory design workshops to envision algorithm tweaks and interface changes. Engineers would build required components and monitor the system. Users would provide perspectives to ensure changes align with their values and needs.

To evaluate intervention effectiveness, we would employ a two-phase process: qualitative methods in the local design context, followed by large-scale A/B testing for system-wide verification. This approach ensures locally observed positive effects translate to the system level. By continuously revisiting the contextual model and indicators based on these evaluations, the platform could incrementally adapt its AI systems to support multidimensional wellbeing objectives. This balanced set of global and local metrics enables holistic progress tracking, potentially moving the platform beyond user satisfaction toward genuine wellbeing alignment.

Having discussed this hypothetical example, we will now proceed to our three real-world case studies.

### Multi-case study

Three design students applied the positive AI design method for their master graduation projects at *Delft University of Technology*. They redesigned or build upon (parts) of existing ubiquitous AI systems to support wellbeing. These redesigns varied in the intervention level (from UI interventions to suggestions for changes in the algorithm) and consequently their impact on wellbeing. None of the students had experience in designing AI or in designing for wellbeing.

Student 1 chose to work in the context of dating apps and was specifically interested in how these could optimize for other components of human identity beyond physical aesthetics. For example, dating apps have hidden mechanics that prioritize physical appearance in their matching algorithms (Parisi and Comunello, 2020; Klincewicz et al., 2022). She wanted to explore how they could also factor in and foster other aspects of identity.

Student 2 chose the context of nutritional and food apps that, for example, track calorie intake and suggest recipes. Such apps tend to prioritize nutritional intake as a proxy for wellbeing. However, an excessive focus on nutritional intake can negatively impact wellbeing, as it often overlooks other aspects of eating that may actually enhance it (König et al., 2021). Therefore, she aimed to broaden their scope to also account for the social and emotional aspects of eating.

Student 3 chose the context of music streaming platforms. The AI recommendation engines in such platforms tend to provide the user with "more of the same" based on listening history and patterns (Tommasel et al., 2022). However, music has powerful potential to influence one's personality, functioning, and understanding of the world. She hoped to leverage the existing AI in such a platform to encourage personal growth and exploration beyond repetitive patterns.

The goal of the multi-case study is to assess two aspects of the design process itself. First, it examines the **efficacy** of the method, looking at whether the designers demonstrate thoughtful understanding of how their decisions potentially impact wellbeing. Specifically, does the method successfully elicit the desired focus on wellbeing considerations from designers, rather than other behaviors? Second, the study evaluates **usability**[12] aspects of the process, such as avoiding unnecessary detours or delays. This refers to whether designers understand the steps involved and feel confident executing them. In other words, the efficacy assessment examines whether the method shapes designer behaviors as intended, while the usability assessment looks at how easily and efficiently designers can apply the process. The inspiration to use the multi-case study to evaluate the efficacy and usability of the method builds on previous work by Tromp and Hekkert (2016), who validated similar aspects using a comparable protocol.

### Procedure

Three student projects utilizing the positive AI method were initiated over a three-week period (March 2023). Each student received a personalized introduction to the project and was provided with reference materials including an overview document of the method, recommended literature, and the "Positive Design Reference Guide" (Jaramillo et al., 2015) to support their research. When the third student commenced their project, a collaborative kick-off meeting was held to establish a shared understanding of the method, address questions, and align expectations across the projects.

---

[12]The term "usability" is sometimes used synonymously with "efficiency" in the literature. However, the concepts of efficiency, efficacy, and effectiveness are often conflated (Zidane and Olsson, 2017). To avoid confusion, this study uses the term "usability" as it encompasses efficiency and has been used to refer to method efficiency by Cash et al. (2023).

Over the remainder of the project, the students met weekly with the supervisory team for guidance. This structured approach aimed to sufficiently equip the students with the necessary understanding and resources to effectively apply the positive AI method within their individual projects. By providing one-on-one introductions, reference materials, a collaborative kick-off, and ongoing supervision, the aim was to support the students in successfully utilizing the positive AI approach.

Then, to gather information on the method efficacy and usability, multiple sources of information were consulted. These included observations taken during the weekly meetings, progress reports and presentations, the final design outcomes and reports, and three recorded and transcribed one-on-one interviews with the students. The weekly meetings provided a platform for the students to share problems they encountered. Oftentimes, they faced similar hurdles, which were carefully documented. The progress reports and presentations served as useful post hoc data to examine how the students were dealing with challenges, how the process developed, and how the designs developed over time. Finally, the interviews aimed to substantiate key themes identified throughout the project period. The first author analyzed the various data sources and shared the findings with the students to ensure accuracy. Before presenting these results, we provide a brief overview of the final design outcomes.

### Materials: design outcomes

In this section, we briefly summarize the final design concepts resulting from the three student projects applying the positive AI method.[13] We present the core functionality and wellbeing goals addressed by each design to provide context before examining the process evaluation findings.

### MiHue

Student 1 designed a dating app called "MiHue" that leverages AI to enhance users' experience of autonomy and relatedness. The core concept balances the needs for uniqueness (autonomy) and connection (relatedness) by highlighting individuality within similarity.

To identify which wellbeing aspects to focus on, Student 1 began by researching the user's contextual wellbeing needs and experiences in using dating apps. Her literature analysis revealed autonomy and relatedness as salient wellbeing facets impacted by dating platforms. To further refine her contextual understanding of wellbeing, she also held multiple generative workshops with target users. During these co-design activities, participants also ideated potential design interventions focused on supporting self-expression and social bonds.

Synthesizing her findings, Student 1 operationalized autonomy and relatedness within her context as the ability for users to express their unique attributes (autonomy) and to find meaningful connections based on shared interests or experiences (relatedness). She then formulated her design directions aimed at enhancing social connection by highlighting individuality within similarity – also referred to in design literature at "Autonomous Yet Connect" (Blijlevens and Hekkert, 2019). This approach focused on promoting a shared connection through uniqueness and common ground, using the AIxD Ideation cards (Piet and MOBGEN, 2019) to link technology capabilities with desired wellbeing outcomes. This led to new features that encouraged users to share more personal and diverse aspects of their identities, beyond physical appearance. One key feature was an improved profile creation tool that prompted users to respond to creative and introspective (AI-generated) questions, facilitating deeper self-expression. Another feature was an algorithm designed to match individuals based on not only mutual interests but also shared values and life goals, aiming to foster more substantial and meaningful connections.

To closely mirror real-world application, she developed a strategy for implementing the novel features either within an existing app or as a standalone platform. Subsequently, she conducted user testing with an interactive prototype (designed using Figma). These tests included questions based on her earlier operationalization of the contextual wellbeing model such as those related to autonomy (e.g., "How well does the app allow you to express your true self?") and relatedness ("Can you describe any interactions you had through the app that made you feel understood or belonged?"). This process was aimed at gathering feedback to refine the recommendations for subsequent iterations, thereby embodying the method's emphasis on continuous alignment. Her subsequent recommendations for the next phase emphasized expanding the focus to encompass additional aspects of wellbeing not initially covered but highlighted in the theoretical model, such as self-acceptance, positive emotions, and physical health. Moreover, she underscored the importance of including diverse user groups, particularly minorities, and considering gender differences, to ensure a more inclusive and holistic approach to enhancing wellbeing through the app's usage. Figure 3 presents this design in three simple illustrations.

### FoodVibe

Student 2 designed an app called "FoodVibe" that uses AI capabilities including facial recognition, natural language processing, and machine learning to provide personalized recipe and dining recommendations tailored to users' specific social contexts and past preferences. The adaptive system aims to promote wellbeing through home dining by encouraging mindful eating and nurturing social connections with shared meals. The idea for FoodVibe originated from the fact that existing nutritional and dieting apps tend to emphasize calorie intake and nutritional intake rather than other aspects of eating that affect wellbeing.

To identify what wellbeing aspects to prioritize in her design, Student 1 conducted a literature study as well as experience sampling (Van Berkel et al., 2017). Her literature analysis revealed mindfulness, social connections, autonomy, and engagement as salient yet overlooked facets. She then combined this information with results from the sampling study to map a user journey specifically aimed at understanding when certain wellbeing experiences may occur. Next, to refine her understanding, she hosted two generative workshops where users emphasized the value of reflection, awareness, and social aspects around meals. This led her to operationalize wellbeing in her context (eating at home) as being present (e.g., engaging with your meal instead of the television) and having a sense of belonging (e.g., feeling related to your family when cooking a nostalgic dish).

This led to the design of FoodVibe which enhances wellbeing by encouraging mindful dining at home, giving users autonomy in their food choices, and deepening connections with dining companions. Utilizing AI, FoodVibe personalizes recipe suggestions by analyzing users' dietary preferences, the people they eat with (identified through facial recognition), and past meal satisfaction. Its features focus on personalizing meal recommendations, enriching social interactions by connecting with friends within the app and aligning meal choices with the group's tastes, and promoting self-reflection on dining experiences to boost wellbeing aspects like autonomy, positive relationships, and mindfulness.

---

[13]The showcases of the projects are available at *anonymized*.

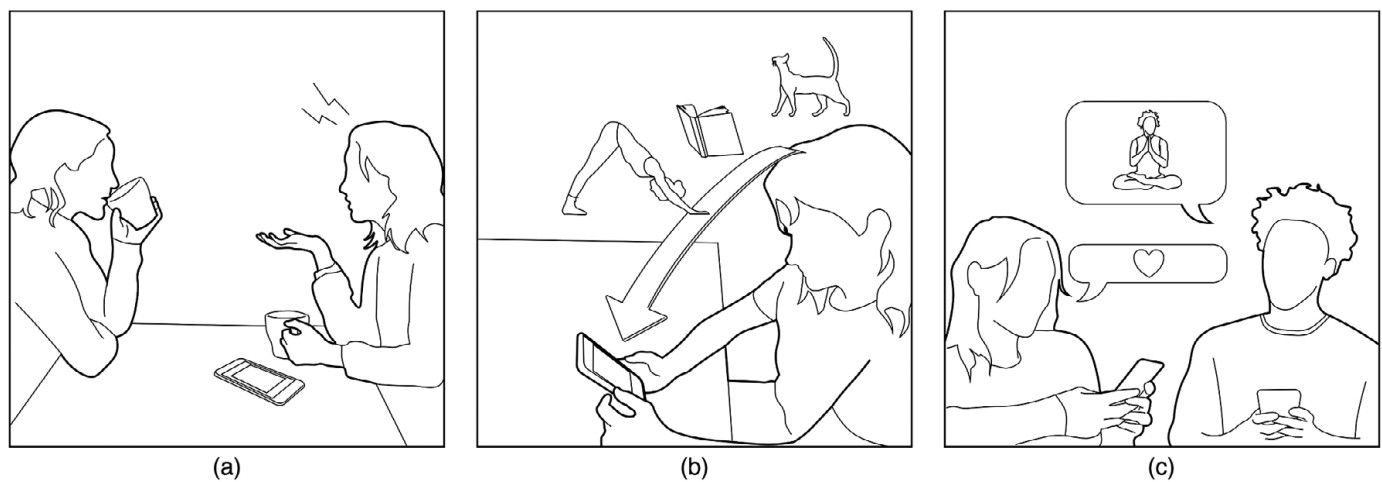

**Figure 3** Three visuals used to illustrate key aspects of MiHue's journey as presented in the expert study: (a) the protagonist's frustration with current dating apps that focus on looks over personality; (b) the protagonist entering their interests during MiHue's enhanced account creation process that encourages authentic self-representation; (c) the protagonist matching with someone who shares common interests, as highlighted by MiHue's features that spotlight unique and shared traits between users to foster meaningful connections.

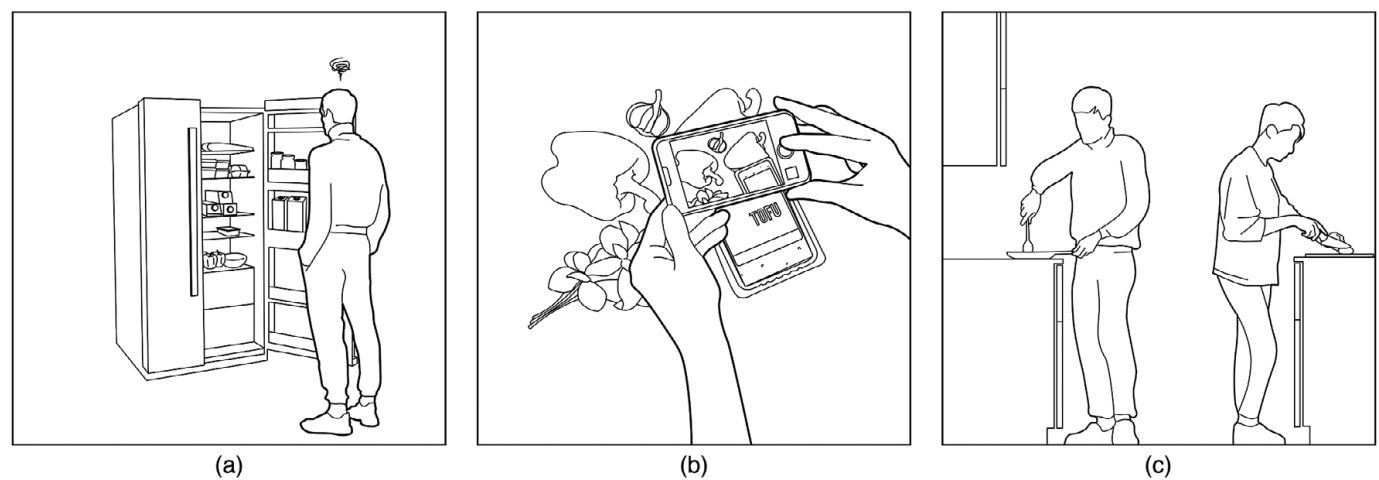

**Figure 4.** Three visuals used to illustrate key aspects of the FoodVibe journey as presented in the expert study: (a) a user frustrated with nutritional limitations when deciding what to cook; (b) the user utilizing FoodVibe's "Recipe Generator" by taking photographs of ingredients on-hand so the app can suggest customized recipes; and (c) two people cooking together in the kitchen, representing FoodVibe's goal of promoting wellbeing through shared meals and human connections.

The final design was evaluated through user testing with a high-fidelity prototype. The evaluation focused on whether the app achieved its design vision and goals, emphasizing the enhancement of eating experiences and perceived wellbeing of users of healthy-eating app. The effectiveness of FoodVibe was assessed based on metrics related to autonomy, positive relationships, mindfulness, engagement, fun, and the overall usability and desirability of the app. These metrics were derived from theoretical models and earlier phases of the research. This led to recommendations for subsequent iterations focusing on user experience improvements, like using avatars for privacy, broadening wellbeing theories to various dining contexts, boosting AI accuracy for tailored recommendations, and conducting thorough user testing for long-term wellbeing impacts.

To illustrate this design, Figure 4 presents it through three simple diagrams.

#### Explore More

Student 3 designed a new *Spotify* feature called "Explore More" that uses the platform's algorithms and extensive music catalog to guide listeners through unfamiliar genres in a personalized way. This idea stems from the observation that existing personalization mechanics, such as "Discover Weekly" playlists, tend to converge on a type of music which over time can get uninspiring. The goal is to expand users' musical tastes and perspectives to foster personal growth and empathy.

She began by analyzing the current landscape of music streaming service features and linked these to wellbeing literature. Her analysis revealed personal development, specifically through music's potential to facilitate self-discovery, as an impactful yet underutilized application for enhancing user wellbeing. Synthesizing her contextual findings, the student recognized limitations of Spotify's existing personalized discovery playlists driven by recommender systems, which can restrict users within narrow musical preferences over time.

Upon recognizing these limitations, she directed her efforts toward intentionally utilizing music's capacity for perspective expansion and self-discovery to promote personal growth (Hallam, 2010). In other words, she operationalized wellbeing as increased engagement with unfamiliar music genres, hypothesizing that exposure diversity would lead to self-development. This inspired the design of Explore More, a feature that could either be integrated into a service like Spotify or as a stand-alone third-party interface. Features included an interactive genre map to visually navigate through unexplored musical territories, guided discovery paths offering sequences of new genres tailored to the user's tastes, personalized recommendations within those genres to ensure a resonant listening experience, self-reflection prompts aimed at deepening users' introspective engagement with the music, and a feedback and adaptation mechanism to refine future explorations based on users' experiences and preferences.

She developed an interactive Figma prototype for user testing, revealing key insights into navigation ease, the effectiveness of discovery paths, and the resonance of music recommendations. Self-reflection prompts were particularly noted for deepening users' personal insights and musical connections. Based on feedback, she recommended refining the UI for better navigation, enhancing the recommendation algorithm for tailored music exploration, and deepening self-reflection prompts for richer introspection. Additionally, she proposed adaptive feedback mechanisms to align the exploration journey with users' changing tastes, ensuring Explore More effectively supports personal growth and musical discovery. The design is illustrated in Figure 5.

### Results of the case studies

To reiterate, a key aspect of evaluating the method is assessing its efficacy by examining whether the designer understands the wellbeing impacts and thoughtfully considers them in their design decisions. Additionally, it involves grasping the relationships between wellbeing dimensions and system components. On the other hand, method usability refers to aspects such as avoiding unnecessary detours or delays, understanding the steps involved, and feeling confident in executing the method Figure 5.

The multi-case study revealed both strengths and weaknesses of the positive AI design method when applied by novice designers.

In terms of method efficacy, students initially struggled to feel confident in their understanding of the wellbeing literature. This was partly due to their unfamiliarity with the field. That is, the breadth of literature was overwhelming, causing uncertainty about when enough research had been done to proceed. This resulted in hesitancy during key stages as students were unsure they grasped concepts well enough to move forward.

*"If you were to do this with other people, you could define the content together and find out, for example, what kinds of things you are missing."* – Student 1

Additionally, the lack of familiarity with translating wellbeing goals into technical requirements or metrics affected designers' ability to thoroughly address wellbeing aims in their solutions. The unfamiliarity with core wellbeing concepts led to doubts about properly executing the methodology. Overall, students lacked confidence evaluating wellbeing considerations throughout the process.

*"You don't really know when you are doing it right."* – Student 2

Initially, designers faced challenges in evaluating and addressing wellbeing in their design process. However, as the project progressed, the methodology compelled them to consistently test their approaches against the specific context and the people involved. This iterative process gradually sharpened their focus and deepened their understanding. By the project's end, this rigorous application resulted in a notable improvement in their ability to integrate wellbeing considerations, as evidenced by their satisfaction with the final solutions. This evolution underscores the methodology's effectiveness in facilitating a contextually relevant approach to wellbeing in design.

In terms of method usability, the process involved iterative transitions between research, ideation, and prototyping rather than a linear sequence. This is often the case for design processes; however, the framing of the method as steps and the initial visualization of them (not included here) gave them the idea it would be more linear. Further, the overall structure provided helpful guidance, but some inefficiencies occurred due to unclear context definition early on and lack of knowledge about wellbeing. Therefore, significant time was spent exploring literature not directly relevant. To some, this may have felt as a waste of time. In reality, it is likely designers will explore directions that in the end of their project may not be relevant anymore – this is what makes them a good designer, being open to multiple avenues and perspectives. The designers suggested to include an introductory wellbeing course, a better explanation of the context early on, and to spend more time planning upfront. Despite inefficiencies, designers were satisfied with their process and outcomes overall.

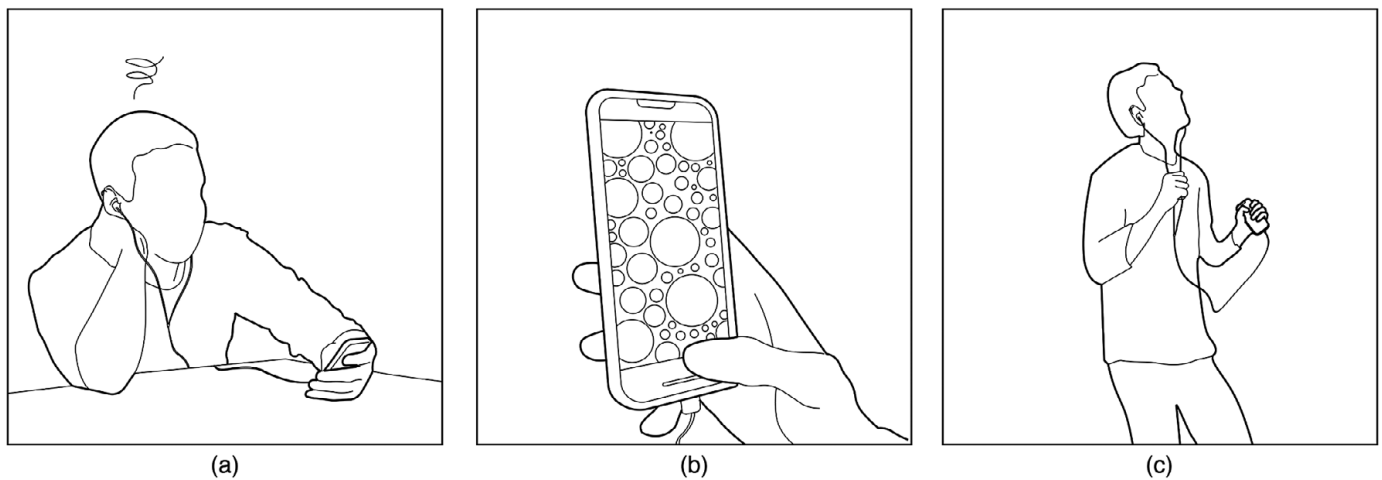

**Figure 5.** Three visuals used to illustrate key aspects of the Explore More journey as presented in the expert study: (a) a bored user unsure what to listen to; (b) an interactive map of music genres that lets users visually browse and see their tastes in context; (c) a user happily dancing after Explore More recommended an unfamiliar yet related genre, demonstrating how it aims to broaden perspectives and facilitate personal growth through personalized music discovery.

In summary, the positive AI method demonstrated efficacy in guiding novice designers to translate high-level wellbeing goals into concrete design proposals grounded in user values. Key steps like contextualizing and iterative development focusing on emergent wellbeing priorities were effective, as evidenced by resulting concepts aligned with community experiences. Despite its inefficiencies, designers expressed satisfaction with the methodology, as it allowed them to address complex tasks through repeated testing against user perspectives, gradually improving contextual understanding. Thus, there is potential to enhance effectiveness in combination with refinements aimed at improving usability.

In terms of method efficacy, key points of attention are as follows:

- Navigating expansive wellbeing literature overwhelmed novices, causing uncertainty grasping concepts to advance confidently. Clearer guidance on scope is required.
- Unfamiliarity translating qualitative aims into technical specifications made operationalization challenging. More extensive scaffolds are needed to aid comprehension.
- Initially lacking familiarity with core wellbeing concepts hampered confidence assessing impacts. But this grew through repeated engagement as understanding increased over time.

Additionally, regarding method usability, certain inefficiencies emerged despite the beneficial structure:

- Unclear initial scope caused detours exploring tangential literature. Signposting priorities earlier would help.
- Absent examples induced difficulty judging step completion. Providing benchmarks would resolve ambiguity.
- Operationalization demands proved taxing for novices. Enhanced support could alleviate strain.
- Shifting between abstract and concrete perspectives around wellbeing felt jarring. Framing this dynamic approach as integral to the design process could smooth transitions.

Having addressed the intricacies of applying the positive AI method through novice designers' experiences and identified areas for refinement to bolster both its efficacy and usability, we now turn our attention to the last research question. In the following section, we present a narrative-based study involving experts to evaluate the quality of the AI system concepts resulting from the application of the positive AI method.

### Narrative-based study with experts

The goal of this study is to assess the design quality of AI systems aimed at enhancing human wellbeing. We chose to use a narrative-based study method, following the example of Tromp and Hekkert (2016) who used this approach to analyze a social design method. Narratives are useful tools for envisioning and assessing the potential impact of emerging technologies that are difficult to prototype or do not yet exist. As Tromp and Hekkert (2016) note, narratives allow people to imagine hypothetical situations as if they were real (Shapiro et al., 2010), providing a means to explore near-future scenarios involving novel technologies (Bleecker, 2022).

By crafting narratives about not-yet-existent AI technologies, experts can then analyze them to evaluate three key dimensions: technical feasibility (Could the required algorithms be developed?), business desirability (Would companies want to develop this?), and outcome plausibility (Could the proposed design plausibly achieve the intended wellbeing benefits?). It is important to consider business incentives when designing AI aimed at promoting wellbeing, since company objectives constrain system behaviors. Without accounting for profit motivations, proposed interventions may conflict with core financial goals.

### Method

#### Procedure and participants

The study involved 17 experts participating in an online questionnaire where they read three narratives describing AI system concepts aimed at enhancing wellbeing. The participants were selected based on their expertise in design, AI, wellbeing, or a combination of these fields. All participants identified as experts in design, seven as experts in wellbeing, and 10 in AI. They were invited to participate via email.

In the questionnaire, participants first read a narrative envisioning a near-future scenario showcasing one of the AI system concepts. After reading each narrative, they answered two comprehension questions about the concept. Participants then completed a 4-item questionnaire assessing the following dimensions on a 7-point Likert scale ("strongly disagree"–"strongly agree"):

1. "The narrative is realistic and believable."
2. "The suggestion that *[AI system]* promotes wellbeing is realistic."
3. "It would be attractive for a company to develop a platform like *[AI system]*."
4. "It would be feasible for a company to develop a platform like *[AI system]*."

This process was repeated for a total of three narratives. The full questionnaire took approximately 15–20 minutes to complete.

#### Materials: narrative development

In crafting the narratives, we followed guidelines discussed in Tromp and Hekkert (2016). The narratives were developed by the authors of this study in collaboration with the students who created the AI system concepts. The ubiquitous contexts of dating apps, food-tracking apps, and music streaming platforms provided plausible scenarios while sidestepping charged assumptions. By carefully considering factors influencing perceived realism when designing the final narratives (700–900 words long), the aim was to elicit unbiased evaluations of the AI concepts and their wellbeing claims. The narratives were illustrated with three graphics each that have also been used to visualize the concepts as shown in the previous section.

A preliminary pilot study (*n* = 5) was conducted to evaluate the realism of the three main narratives using the Perceived Realism Scale (Green, 2004) before proceeding to the main study. Following this initial pilot, no changes were deemed necessary. However, minor updates were made after the narratives were copyedited by a native English speaker. To assess realism in the complete questionnaire, we used a single adapted item: "The narrative is realistic and believable." The full narratives are provided in the Appendix.

#### Results of the narrative-based study

Figure 6 presents a graph of the results from the narrative-based study involving expert evaluations across three distinct concepts:

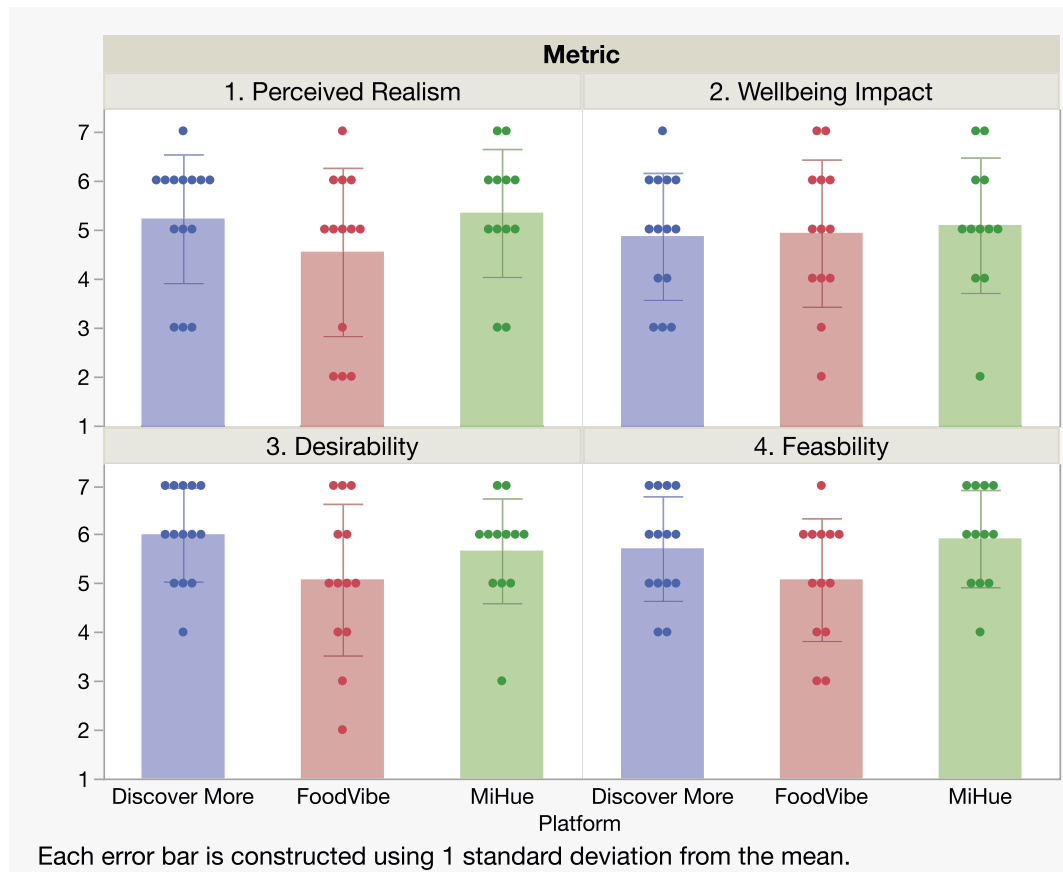

**Figure 6.** Barchart comparison of expert evaluation ratings across the three concepts: MiHue, FoodVibe, and Explore More. Metrics visualized include perceived realism, wellbeing impact, business desirability, and business feasibility.

MiHue, FoodVibe, and Explore More. The concepts, received moderately high mean ratings on perceived realism, impact on wellbeing, business desirability, and business feasibility. The concepts, on average, received moderately high ratings across the metrics. Specifically, Explore More was rated highest in business desirability, while MiHue and FoodVibe showed similar ratings in perceived realism and wellbeing impact, respectively. Notably, there were no significant differences in ratings among the different expert groups. The associated standard deviations indicate a moderate variation in expert opinions. On the qualitative front, feedback from two participants indicated that the narratives were lengthy, while another two experts remarked that the stories leaned toward being overtly positive. However, they also noted their understanding of this positive skew, acknowledging the study's context aiming to portray an ideal user experience. Further insights from the expert's feedback will be addressed in the subsequent sections.

## Discussion

This study introduced the positive AI design method for developing AI systems that actively promote human wellbeing. Following the framework for evaluating design methods proposed by Cash et al. (2023), we have provided a "chain of evidence" through multiple studies to assess the credibility and robustness of the positive AI method.

Specifically, we first discussed the motivation for the method based on gaps in current AI design processes. We then explained the nature of the method as a principle-based approach suited for ubiquitous AI systems that seek to actively integrate wellbeing. Next, we detailed the iterative development process applying a cybernetic framework. We then outlined the key steps: 1) contextualization, 2) operationalization, 3) design, 4) implement, and 5) reiteration. Finally, we presented evidence for the method's impact claims through a multi-case study with novice designers and an expert evaluation assessing the quality of the resulting concepts. Through this initial validation, the method showed promise for improving AI design while also revealing areas needing refinement.

In this final section, we will first briefly reflect on the multi-case and expert studies, after which we will discuss their limitations. Then, we discuss the position of the positive AI method with respect to existing frameworks such as VSD and IEEE-7010. Finally, we outline proposed adaptations and future work based on the outcomes of this discussion.

### Reflections on the case study: comparing efficacy and usability

Our exploration through three case studies illuminated how designers integrated wellbeing into AI design, effectively addressing *RQ2*, by uncovering the method's strengths and weaknesses. Specifically, the study revealed a trade-off between the method's efficacy and usability. While the process successfully directed the students' focus toward considerations of wellbeing impacts in their design concepts, demonstrating the method's efficacy, the students also experienced inefficient detours and uncertainty during certain steps of the process. This indicated usability challenges with the current version of the method. For example, the students reported being overwhelmed by the breadth of literature when researching wellbeing theory, causing uncertainty about when sufficient research had been conducted to move forward. Such detours revealed usability issues despite the method's efficacy in eliciting wellbeing considerations. To extend this, Andreasen (2003) emphasizes designers must develop proper mindsets, not just learn procedures, to effectively utilize methods. Future iterations of the method could incorporate visual models and clearer explanations of theory to aid in building this effective mindset, potentially mitigating some of the observed usability issues.

While discussions of usability and efficacy trade-offs are relatively absent in academic literature on design methods, both factors are critical drivers of tool adoption in practice. Farzaneh and Neuner (2019) argue for the indispensability of usability in the effective employment of tools, a point nuanced by Eason (1984), who notes that heightened functionality may inadvertently compromise usability. This insight suggests that while poor usability can significantly hinder adoption, it does not completely negate the potential of effective tools. Supporting this notion, Nielsen (1994) highlights that a method's acceptance hinges on both its utility (efficacy) and its ease of use (usability). Consequently, evaluating both of these aspects will be crucial for improving methods and conducting future research (Blessing and Chakrabarti, 2009).

### *Reflections on the expert study: assessing impact, desirability, and feasibility*

The expert evaluation study aimed to assess the quality of the AI system concepts resulting from application of the positive AI method, providing an answer to the third research question (RQ3). Beyond just examining the intended wellbeing impacts, quality was operationalized more broadly through three key dimensions: technical feasibility, business desirability, and outcome plausibility.

While the positive AI method specifically focuses on supporting wellbeing considerations, results indicated that the concepts also scored moderately high on feasibility and desirability. If the method only enabled wellbeing aims, we may have expected substantially lower ratings on the other dimensions. For example, redesigning a platform like *Instagram* for maximum wellbeing benefit may result in features that lock out users for a given amount of time to stimulate physical exercise instead. This would, obviously, not be desirable from a business perspective. Notably, the concepts were not rated perfectly high across all metrics. That is, when inspecting results closely, variation existed both within and across the projects, indicating varying expert perspectives rather than uniform positivity.[14] This variety in feedback underscores the validity of our data and demonstrates that the method can elicit a diversity of responses, highlighting its effectiveness and the nuanced nature of integrating wellbeing into AI design.

Nonetheless, the overall promising ratings for technical feasibility, business desirability, and outcome plausibility imply that the method provides useful scaffolding for creating AI concepts aligned with multiple stakeholder needs. This suggests that the method's emphasis on contextualization and continuous alignment effectively supports the consideration of multiple stakeholder perspectives, including those of users, businesses, and technical teams. However, it is important to approach these outcomes with caution. The controlled environment of the study may not fully replicate the complexities encountered in real-world applications, necessitating further empirical investigations to validate the method's applicability. Additionally, a relevant question arises: Would students provided only a general prompt to design for wellbeing, without the structured positive AI method, create concepts of significantly lower quality in terms of feasibility, desirability, and plausibility? A future study comparing outcomes from students given just a design prompt versus those using the full positive AI method could provide insights into how specifically the method enhances these quality dimensions beyond the impact on wellbeing. Both the multi-case study and expert evaluation successfully identified strengths and areas for improvement. However, these methods have certain limitations, which we will discuss in the following section.

### *Limitations of the study*

While our research approach provided valuable insights, it is important to consider its implications for future work needed to further validate and refine the positive AI method. First, the student projects provided useful initial insights into the application of the positive AI method by novice designers, as they were able to engage with the method in-depth rather than superficially, helping surface the issues discussed in this article. However, it is important to note that the students received regular coaching and guidance from the supervisory team who created the method. This support likely influenced their ability to apply the method successfully and mitigated some of the challenges they faced, such as navigating the breadth of wellbeing literature or translating abstract concepts into concrete metrics.

Moreover, the academic setting differed from professional design environments where designers are presumed to possess extensive domain experience. The temporary nature of the student projects made it difficult to evaluate the long-term, sustained application of the positive AI approach over multiple iterative cycles. The compressed student timelines allowed only for a single, but thorough, pass through the steps, precluding the examination of how designer understanding and execution might improve through repeated application over many months or years and limiting the analysis of how contextual models develop over time as relationships and priorities shift.

Furthermore, it remains to be seen how the method would be applied by a group of computer scientists who may or may not have a designer on the team and lack access to the authors' expertise. Future work should explore how successful the method is when design teams do not receive expert guidance, possibly by comparing the outcomes of teams given the method with varying levels of training and support. This could provide valuable insights into the method's effectiveness and usability in more realistic scenarios.

To fully assess the long-term, repeated application of the positive AI method, in-depth research in professional settings is necessary. This would involve tracking multiple iterative cycles over extended time periods, providing insights into how designer understanding and execution might improve through repeated application, and how contextual models develop as relationships and priorities shift. Real-world validation by professional designers in industry contexts would provide invaluable feedback on limitations and areas for improvement when applied in practice. Additionally, some parts of the method can be developed further to provide additional guidance, tools, and examples that make the framework more accessible for practitioners. Moving forward, addressing these limitations through real-world validation, additional resources, and studies on longitudinal effectiveness will be crucial. The following sections will discuss how this relates to existing approaches and propose adaptations for the method.

---

[14]Data available upon request through the corresponding author.

### The method and existing approaches

This research was conducted within the broader context of AI design, a field often challenged by the proliferation of frameworks without sustained dialogue. Therefore, in this section, our goal is to connect our method with existing approaches, thereby advancing the field and identifying opportunities for future work.

The positive AI design method addresses a significant limitation within VSD: the absence of mechanisms for assessing how well values are realized in technology design and the tangible impact of AI systems. This challenge, highlighted by Sadek et al. (2023c), is met by embedding wellbeing assessment as a fundamental principle within our approach. By prioritizing wellbeing, our method enhances VSD, making the integration of values within AI systems both measurable and actionable. This focus ensures that abstract values are operationalized into practical design and evaluation criteria, providing a holistic, empirical basis for optimizing human values. This aligns with the argument presented by Harris (2010) that enhancing wellbeing inherently promotes all human values insofar as they empirically contribute to flourishing.

Similarly, the positive AI design method extends and complements the IEEE-7010 standard (Schiff et al., 2020). While IEEE-7010 lays a foundation for the integration of well-being metrics into the life cycle of AI systems, our method takes a step further by directly mapping these wellbeing considerations onto existing design approaches. This direct integration ensures that wellbeing is not only assessed as an outcome but actively shapes AI development from the outset. Furthermore, our method extends the IEEE-7010 framework by offering detailed, practical guidance on mapping wellbeing metrics directly to design decisions, thus facilitating a more granular and actionable approach to enhancing wellbeing through AI systems. This approach not only adheres to the holistic perspective advocated by IEEE-7010 but also advances its application by providing a structured method for translating wellbeing principles into concrete design practices.

Further, this work can be seen as contributing to the broader question of AI alignment, a field primarily concerned with aligning AI technologies with human values (Christian, 2020; Gabriel, 2020). Without going to deep into this area, we recognize an opportunity to advance the field through the methodology presented here. Specifically, the positive AI method takes a human-centered approach that contrasts with some common perspectives in the field of AI alignment. Much alignment research focuses on technical solutions like reward modeling (Christiano et al., 2017; Bai et al., 2022) and meaningful human control (Cavalcante Siebert et al., 2023). These techniques aim to formally specify values and control objectives that AI systems should optimize for. In contrast, the positive AI method emphasizes building contextualized understanding of users through participatory research and design processes. It focuses on continuously aligning systems with the multifaceted and emergent nature of human wellbeing through collaboration. In this way, the positive AI method diverges from alignment approaches that prioritize formal specification of abstract values and control objectives over participatory human-centered design processes. The proposed positive AI method provides a complementary human-centered perspective to balance the prevalent technical focus in this field by enabling alignment techniques to be *sensitive to human experience.* Still, greater synergy is needed between these approaches to ensure both human values and technical reliability are embedded in mutually reinforcing ways. The positive AI method's human-centered approach could be enhanced by integrating formal techniques like reward modeling, which may help scale contextual findings.

Lastly, perspectives from explainable AI (XAI) could enhance the positive AI method. By making transparent how algorithms and data shape user experiences, we can better understand relationships tied to wellbeing (Ehsan et al., 2021). In turn, positive AI's emphasis on establishing causal links between system components and outcomes can progressively demystify the AI system. This increased transparency aids designers in identifying failure points, unintended consequences (Gunning et al., 2019), and can enhance designers' capabilities to co-create with AI (Zhu et al., 2018a). Furthermore, XAI's focus on addressing diverse stakeholder needs is in harmony with positive AI, offering techniques to elucidate system behaviors and supporting the creation of AI that nurtures human flourishing (Felzmann et al., 2020; Larsson and Heintz, 2020). An exemplary instance of leveraging AI explanations to enhance wellbeing is demonstrated by Hume AI's EVI, which not only identifies the emotions present in the user's voice but also clearly indicates the emotions it utilizes for its responses. This approach enriches wellbeing by encouraging deeper, more empathetic communication.

### Proposed adaptations and opportunities for future work

Reflecting on our evaluation of the positive AI method, we recognize its significant contributions toward bridging existing gaps in the field, alongside areas that remain open for improvement. In light of these insights, we propose several avenues to move the method forward.

First, providing designers with examples and heuristics may improve method usability. For instance, developing a framework to determine when enough contextual research has been conducted could prevent unnecessary detours. This framework might involve checklists of key relationships or suggested timeboxes. Likewise, heuristics and examples could increase confidence during overwhelming stages such as conceptualizing and operationalizing wellbeing. A recent study by Peters (2023) synthesizes over 30 years of psychology research to provide 15 of such heuristics that may help technology designers create more wellbeing-supportive user experiences by identifying key areas where AI can significantly impact users' psychological wellbeing, ensuring that designs are grounded in well-established principles. Additionally, fast-paced, preliminary simulations, such as workshops, could acclimate designers to the method and underlying wellbeing theories, offering a practical glimpse into the process and expected outcomes. This would also fulfill a key recommendation from the literature to promote AI education, ensuring positive impacts and encouraging cross-disciplinary collaboration (Schiff et al., 2020; Morley et al., 2021; Bentvelzen et al., 2022).

Furthermore, the current positive AI method focuses primarily on the design phase of the AI life cycle, a conscious choice made to better scope this study. Consequently, it does not yet provide guidance for integrating wellbeing principles across all seven stages of the AI development process. Future research should explore expanding the method's scope and applicability to address this limitation. Extending the method across the entire AI life cycle would ensure consistent application of wellbeing objectives, addressing emerging challenges and unintended consequences (Norman and Stappers, 2015).

Equally important is the investigation of the method's implementation across diverse team compositions. While the method thrives with multidisciplinary teams including human-centered designers, ethicists, and subject matter experts, many organizations may lack such diverse expertise. Future work should explore various team arrangements – from tightly integrated multidisciplinary groups to looser collaborations with external experts – and identify core skills required for effective implementation. This could inform targeted training programs for professionals applying these principles without full specialist support. Addressing both the full AI life cycle integration and team composition challenges is essential for improving the method's real-world feasibility, effectiveness, and broader impact across diverse organizational contexts.

In the same vein, the positive AI method currently does not fully support the activity of co-designing. While we have identified this as an important gap in the literature and have emphasized co-designing, such as the inclusion of stakeholders, as essential to the success of positive AI, we do not actively discuss how this can be best achieved. Due to the scope and length of this article, this is not the most appropriate place for an in-depth exploration of co-design techniques. Perhaps a future platform detailing the method further could include such resources. For now, we recommend referring to the work of Sanders and Stappers (2008) for general guidance on co-design, and other resources more specifically applied to AI (e.g., Liao and Muller, 2019; Subramonyam et al., 2021; Zytko et al., 2022; Sadek et al., 2023a). Future research could investigate how these frameworks may complement each other and enhance the positive AI method.

It is important to acknowledge that while the positive AI method holds promise, its practical implementation poses significant challenges. These include the difficulty of consistently engaging diverse user communities, the resource-intensive nature of continuous participatory design, and the complexities around grasping the concept of wellbeing. Additionally, aligning the positive AI method with existing organizational goals and workflows can be challenging, as it requires cultural shifts and buy-in from multiple stakeholders. As mentioned above, educating AI practitioners about thick concepts such as wellbeing has been suggested as a way of dealing with some of these challenges (Alexandrova and Fabian, 2022); however, the other practical challenges, especially around aligning existing organizational goals and workflows with positive AI, remain open.

Addressing these practical challenges is critical for the method's real-world applicability and effectiveness. Therefore, future research should focus on developing scalable, efficient techniques for stakeholder engagement, refining operationalization processes, and creating adaptable frameworks that can integrate seamlessly with different organizational contexts. Investigating these aspects further will enhance the feasibility and impact of the positive AI method in diverse real-world settings. Investigating efficient techniques for community collaboration at scale, such as those proposed by Peters et al. (2023), would strengthen this vital feedback loop and improve the method's real-world feasibility. Developing more streamlined and scalable approaches to co-design will help the positive AI method better incorporate diverse perspectives and align with the needs and values of the communities it serves.

## Conclusion

The positive AI design method presents a structured approach for developing AI systems that promote human wellbeing. Our evaluation through case studies and expert assessment reveals its potential to improve AI design, while also highlighting areas for refinement. By emphasizing continuous wellbeing assessment and providing practical guidance, this method addresses critical gaps in existing approaches. Future work should focus on enhancing usability, expanding applicability across the full AI life cycle, and developing efficient techniques for stakeholder engagement. As AI becomes an integral part of our lives, methodologies like this will be vital for ensuring that powerful, emerging technologies are not only aligned with human values but also actively contribute to our wellbeing.

**Acknowledgements.** We thank Danielle Klomp, Gigi Wang, and Eve Liu for their contributions to this study, Xavière van Rooyen for the illustrations, and Haian Xue and Stefan van de Geer for their support as coaches.

**Data availability statement.** The datasets generated and/or analyzed during the current study are available from the corresponding author on reasonable request.

**Competing interest.** On behalf of all authors, the corresponding author states that there is no conflict of interest.

**Ethics statement.** The studies involving human participants were reviewed and approved by Human Research Ethics Committee of *TU Delft*. The participants provided their written informed consent to participate in this study.

# Appendix

# Narratives for Expert Study

## A.1.1. MiHue

### A.1.1.1. Sarah Finds Her Person on MiHue

Sarah was a 24-year-old marketing assistant who had recently moved to Amsterdam for her job. Though doing well at work, Sarah hoped to expand her social circle and meet a romantic partner.

Sarah sighed as she swiped left on another dating profile. "No luck tonight?" her friend Amanda asked, noticing her frustration. "Ugh, no," Sarah replied. "I'm so over these apps only focusing on looks and generic interests. The conversations are meaningless." Amanda nodded, "It's impossible to make real connections on them."

"Exactly!" Sarah said. "I want someone I can have deep talks with, not just small talk." Then, Amanda mentioned a new app called MiHue that matched based on compatibility, not appearances. Intrigued, Sarah downloaded it, hoping to find someone she could truly connect with.

Sarah spent time customizing her profile to accurately convey herself as a person. The app first asked for basic information like her age, location, interests, and hobbies. Sarah entered details such as her love of books, yoga, piano music, cooking, and indie films. There was also a section to enter personality traits and values. After thinking about it, Sarah chose words like "kind," "quirky," "adventurous," and "curious." She hoped that showing these parts of her real, authentic self would help attract like-minded matches.

Next was photograph selection. MiHue automatically sorted Sarah's camera roll into categories based on interests she had entered, like "book club," "yoga poses," and even a category for her cats! The app recommended choosing a thoughtful balance of photographs, showcasing interests she shared with many others, as well as unique photographs that highlighted her individuality. Following this advice, Sarah picked a mix of photographs showing herself reading at book club, doing yoga poses, snuggling with her cats, dressed up silly for a costume party, and exploring street markets while traveling solo. She appreciated MiHue's guidance in thoughtfully selecting photographs to give a real glimpse into her life.

Before completing her profile, MiHue generated bio suggestions based on Sarah's selected interests and personality traits. Sarah was pleased to see MiHue recommend phrases and descriptions that she identified with, like "eager world traveler" and "loves learning." After adding this personalized text to her bio, Sarah felt confident she had shown a genuine, multi-sided portrayal of herself. She hoped this openness would attract partners interested in the same kind of real connection.

Sarah then applied filters to further customize the profiles she would see on her swiping screen. She highlighted interests and values important to her, like "book lover," "yoga fan,", "stargazing," and "kindness." MiHue suggested more detailed selections based on Sarah's existing choices, such as her favorite book type, yoga style, and specific constellations. Adding these helped fine-tune her results beyond surface-level interest matches. Sarah was eager to start swiping and see whether these filters would find her perfect partner!

Sarah was excited to see MiHue highlight keywords and interests she had in common with each potential match as she swiped through profiles. Whenever she matched with someone, a pop-up would alert her to any especially unique interests that she and her match shared. Seeing that a match was equally passionate about an obscure fantasy novel series, or appreciated her favorite niche yoga philosophy, immediately captured Sarah's attention. It sparked a feeling of kinship, as these rare commonalities carried more weight and fostered a deeper connection. Sarah realized that even such simple similarities meant much more to her than merely finding someone attractive.

After swiping for quite some time, MiHue checked in to gather Sarah's feedback on her experience so far. Sarah noted how much she appreciated connecting based on shared values, passions, and personality traits, rather than just appearances. MiHue processed this input, and Sarah soon noticed refined highlight suggestions based on the types of profiles she responded well to. With these tweaks, her results improved drastically, saving Sarah endless swiping by discovering ideal matches sooner.

Before long, Sarah matched with David, who shared her love of books, stargazing, cooking, and costume parties. MiHue immediately suggested personalized conversation starters about their favorite constellations and stargazing spots. Sarah felt relieved that the app provided these tailored opening messages, reducing the pressure and anxiety she typically felt when having to make the first move. As she and David continued chatting with MiHue's assistance, Sarah was struck by how smoothly the conversation flowed. Rather than the typical small talk she was used to, they dove into discussing childhood memories, future dreams, and the stresses of moving to a new city.

Overall, Sarah was really impressed with how the MiHue app worked. It helped her show her true self and then actually matched her with people who shared deeper compatibility, not just superficial interests. The app seemed to really "get" her personality and what she was looking for, based on how she filled out her profile and reacted to different matches. She felt like MiHue was tailoring its recommendations just for her, suggesting people who she could build a unique connection with, instead of the usual t generic matches.

After countless disappointing dating app experiences, MiHue had given Sarah renewed hope around finding not just a partner, but someone who would value every part of who she was. Sarah looked forward to building this new meaningful connection with David, and seeing where it led organically without any pressure. She was grateful to MiHue for restoring her faith in the process of open, authentic human connection.

### A.1.2. FoodVibe

#### A.1.2.1. Sascha Explores New Musical Horizons

Sascha had been using Spotify for years, but recently they felt like they were stuck in a musical rut. Playlists like Discover Weekly and Release Radar were starting to feel boring, only suggesting songs in the genres they usually listened to, like pop, indie rock, and folk. Sascha wanted to expand their musical tastes and try new types of music, but every time they tried searching Spotify's huge catalog on their own, they felt overwhelmed and went back to their musical comfort zone.

Sascha wished Spotify had a way to guide them through new genres, encouraging them to try different music while keeping the exploration manageable and curated. One day, while using the app, an ad for a new feature called "Discover More" caught their attention. The description said this interactive experience could introduce listeners to unfamiliar genres in a way tailored to their current listening habits. It promised the chance to gain new perspectives and foster personal growth through the musical journey. Intrigued and inspired, Sascha tapped the big "Let's Go!" button right away.

The app screen changed to show a map, filled with bubbles of all sizes, each representing a different music genre. Some Sascha recognized, while others sounded completely unfamiliar. In the very center pulsed their profile bubble, showing their top genres of indie pop, folk rock, and neo soul. Using the easy touch controls, Sascha zoomed out to see genres spreading across the whole map. They felt excited to explore this new world of music outside their usual tastes.

Guided by Spotify's algorithms, Sascha started moving their profile bubble toward a nearby group of genres they knew about but rarely intentionally listened to country, folk, bluegrass, and Americana. As they approached, the app automatically made a preview playlist mixing popular songs and lesser-known tunes. The twangy vocals, fast banjo strumming, and lyrics about small towns and country life captivated Sascha immediately. They smiled, tapping the heart icon to save several songs to a new playlist appropriately called "Country Roads."

After an hour exploring those genres, Sascha was surprised they had built a country playlist with over 50 songs. It satisfied them in a way they did not expect, making them think about lyrics exploring topics like family, faith, and rural working class life. Occasionally, thought-provoking questions from Spotify showed up on the right side of the screen, like "What emotions do you feel from this music?" and "How might these songs connect you to new people or places?" Sascha liked that these prompts helped them to reflect on how the music impacted their feelings and views.

Ready for the next part of their journey, Sascha used the touch controls again to zoom out and browse nearby areas. One cluster labeled Afrobeat, reggae/dub, soca, and dancehall caught their attention. They moved their profile bubble there, excited by the preview's lively instruments, upbeat rhythms, and chanting vocals. As the first few songs played, Sascha's shoulders started swaying instinctively to the infectious beats. The music felt vibrant, celebratory, and liberating. They soaked in information about each genre's history while listening, appreciating them in a richer context.

After a while, Sascha glanced at the map and was amazed to see how far their profile bubble had moved from the center. Music styles they never would have tried before now characterized their soundscape. Sascha realized this journey had expanded their tastes in ways they did not think were possible, unlocking new understandings and perspectives.

As the hours passed by quickly, Sascha felt themself growing mentally tired. But they were thrilled by all the new music worlds they had uncovered. Looking at their library, they now had playlists labeled Country Roads, Island Vibes,

African Beats, and more. It was time to finish this session, but Sascha knew this was just the beginning of a lifelong musical adventure. They could return to Discover More anytime, choose a new direction, and keep growing.

Thinking about the experience, Sascha was grateful to Spotify for making Discover More. Far from just an algorithm-driven music finder, it felt like a service designed to broaden Sascha's perspectives while respecting their choices. The app had achieved its goal: personal growth through exploring music. Sascha went to bed that night feeling their world had expanded, with endless possibilities ahead.

### A.1.3. Explore More

#### A.1.3.1. Andrea's Path to Mindful Eating

It was around 6pm when Andrea's stomach started growling. They needed to figure out dinner. With a groan, Andrea went to the kitchen. Opening the fridge, they saw some vegetables, tofu, yogurt and condiments. The cabinet had a few canned goods and some pasta. Lately, their life had felt very busy with work and friends. Making dinner was often forgotten – most nights they would just order takeout food or heat up a frozen meal.

But for some reason, Andrea did not feel like more greasy takeout tonight. They wanted something homemade and healthy. If only they had more ingredients to use…

Suddenly, Andrea remembered their friend Taylor mentioning a meal planning app called FoodVibe. "It's great! It helps me cook more carefully and be mindful about my eating habits," Taylor had said excitedly. "You have to try it, Andrea!" Well, now was a good time, Andrea thought. They downloaded FoodVibe on their phone and made a profile. For their goal, they put "eating more carefully."

Andrea found the app very easy to use. It immediately asked them to take some photographs of the ingredients they had. Andrea arranged the vegetables and tofu nicely for a little photoshoot and then uploaded the pics to the app. In a matter of seconds, FoodVibe made a list of recipes they could make using just those ingredients. One dish, a vegetable stir fry, looked good to Andrea – perfect for tonight!

As they started preparing, Andrea tried to follow the careful eating advice from FoodVibe. They focused on the colors and textures of the vegetables as they chopped…the sizzling sounds as the food hit the pan…the delicious smells filling the kitchen. Cooking this way felt calming, almost meditative. Before Andrea knew it, their stir fry was done! They quickly took a pic for the app before eating.

The meal tasted amazing – fresh, healthy, and so satisfying. Andrea felt proud that they made it themselves with just the ingredients they had. After eating, they labeled the photograph in FoodVibe as "tasty" and "easy" and saved it to look at later.

Over the next few weeks, Andrea used FoodVibe daily to plan and log meals. Taking food photographs and labeling recipes became a helpful routine, creating a visual record that made them appreciate and think about each meal more. Looking at their FoodVibe journal also gave Andrea some important insights. They saw that although takeout had made up most of their diet, cooking healthy meals at home gave them energy in a different way.

Using the app's features regularly helped Andrea get more organized with preparing food. They learned go-to homemade recipes they loved eating again, including that tasty vegetable stir fry. Following FoodVibe's careful eating advice improved Andrea's enjoyment of home cooking. Over time, using FoodVibe gave Andrea a real sense of achievement – they were making real progress toward their goal of developing a healthier relationship with food. The app provided helpful tools that supported their continued learning and growth around careful eating.

A few weeks later, Andrea met their friend Taylor for coffee. Andrea and Taylor had been close since college, but recently they had not been seeing each other that often due to their busy lives.

"Thanks for telling me about FoodVibe – I'm loving it!" Andrea said.

"So happy it's working for you as well! We'll have to get together and cook something fun from it soon." Taylor suggested.

"That's a great idea! Let's plan a dinner date." Andrea replied excitedly.

They both opened the FoodVibe app on their phones. Andrea and Taylor tapped on the "Food Friend Finder" feature. Suddenly, each of their apps detected that another user was sitting close by. Based on the overlap of recipes they had logged as having enjoyed, FoodVibe recommended a Mediterranean chickpea skillet for their dinner date.

"Ooh that looks delicious, let's make it together!" said Taylor.

Andrea smiled, excited to reconnect more with their old friend over a home-cooked FoodVibe meal. They were grateful the app could bring users together in such a tangible way.

The day of their dinner date finally arrived, and Andrea went over to Taylor's apartment, excited to cook with their friend again. In the kitchen, they scrolled through the Mediterranean chickpea skillet recipe in the FoodVibe app, splitting up the tasks.

Once the skillet was in the oven, the two friends caught up on life while sipping wine. It felt just like old times. When the timer went off, they opened the oven to reveal a beautifully aromatic dish.

Over the meal, Andrea and Taylor continued bonding over their love of food. They took pictures of the delicious chickpea skillet to log in the app later. Andrea labeled the dish as "fun," "exciting," and "easy" in FoodVibe. They knew the app could use these labels to recommend similar fun and easy recipes to make in the future. After dinner, the pair browsed FoodVibe some more, planning more recipes to cook the next time they got together.

Andrea felt so grateful for the app bringing them and Taylor back together. They hoped that they would keep using FoodVibe to explore mindful cooking and reconnect over homemade food. The app provided an easy way to share recipes, photographs, and memories.

In the following weeks, Andrea and Taylor met up to cook several more times. They loved learning new recipes, cooking tips, and nutrition facts together through FoodVibe. Using FoodVibe became a regular ritual that strengthened their friendship, helping them form deeper bonds with each other through food.

**Positive AI Design Method – Checklist for Developing AI to Enhance Wellbeing**

*Contextualization phase:*

- Review relevant wellbeing literature and theory
- Map key components of the AI system (algorithms, interface design, etc.)
- Conduct qualitative user research (interviews, focus groups, etc.)
- Synthesize theoretical and user research findings into a contextual wellbeing model

*Operationalization phase:*

- Select validated global wellbeing scales
- Develop context-specific wellbeing metrics linked to system components
- Ensure metrics enable optimizing algorithms to enhance wellbeing

*Design phase:*

- Identify high-potential targets for design interventions via surveys, behavior tracking, etc.
- Envision system modifications across layers (UX, algorithms) to impact wellbeing
- Produce artifacts like journeys maps and design principles encoding wellbeing aims

*Implementation phase:*

- Guide development process using artifacts from design phase
- Ensure implemented features align with envisioned optimized interactions

### A.1.4. Continuous alignment

- Regularly re-engage user community via interviews, workshops, etc.
- Revisit contextual model to realign priorities
- Repeat full process to incrementally enhance wellbeing impacts